%% file: root.tex
\documentclass[a4paper, 10 pt, conference]{ieeeconf}  

\IEEEoverridecommandlockouts                              

\usepackage{graphicx} 
\usepackage{siunitx}
\usepackage{multicol}

\usepackage[utf8]{inputenc}
\usepackage{pgfplots}
\DeclareUnicodeCharacter{2212}{−}
\usepgfplotslibrary{groupplots,dateplot}
\usetikzlibrary{patterns,shapes.arrows}
\pgfplotsset{compat=newest}

\usepackage[percent]{overpic}

\title{\LARGE \bf
Exploring the Capabilities and Limits of 3D Monocular Object Detection - A Study on Simulation and Real World Data
}

\author{Felix Nobis$^{1}$, Fabian Brunhuber$^{2}$, Simon Janssen$^{2}$, Johannes Betz$^{1}$ and Markus Lienkamp$^{1}$
\thanks{$^{1}$Felix Nobis (corresponding author), Johannes Betz and Markus Lienkamp are with the Chair of Automotive Technology, Technical University of Munich {\tt\small nobis@ftm.mw.tum.de}}%
\thanks{$^{2}$ Fabian Brunhuber, Simon Janssen are master students at the  Technical University of Munich} 
}

\usepackage{textcomp}
\usepackage{lipsum}
\newcommand\copyrighttext{%
    \footnotesize \textcopyright 2020 IEEE.  Personal use of this material is permitted.  Permission from IEEE must be obtained for all other uses, in any current or future media, including reprinting/republishing this material for advertising or promotional purposes, creating new collective works, for resale or redistribution to servers or lists, or reuse of any copyrighted component of this work in other works.
}
\newcommand\copyrightnotice{%
    \begin{tikzpicture}[remember picture,overlay]
    \node[anchor=south,yshift=10pt, xshift=10pt] at (current page.south) {\fbox{\parbox{\dimexpr\textwidth-\fboxsep-\fboxrule\relax}{\copyrighttext}}};
    \end{tikzpicture}%
}

\begin{document}

\maketitle
\copyrightnotice
\thispagestyle{empty}
\pagestyle{empty}

\begin{abstract}
3D object detection based on monocular camera data is a key enabler for autonomous driving. The task however, is ill-posed due to lack of depth information in 2D images. Recent deep learning methods show promising results to recover depth information from single images by learning priors about the environment. Several competing strategies tackle this problem. In addition to the network design, the major difference of these competing approaches lies in using a supervised or self-supervised optimization loss function, which require different data and ground truth information. In this paper, we evaluate the performance of a 3D object detection pipeline which is parameterizable with different depth estimation configurations. We implement a simple distance calculation approach based on camera intrinsics and 2D bounding box size, a self-supervised, and a supervised learning approach for depth estimation.

Ground truth depth information cannot be recorded reliable in real world scenarios. This shifts our training focus to simulation data. In simulation, labeling and ground truth generation can be automatized. We evaluate the detection pipeline on simulator data and a real world sequence from an autonomous vehicle on a race track. The benefit of simulation training to real world application is investigated. Advantages and drawbacks of the different depth estimation strategies are discussed.
\end{abstract}

\section{Introduction}
Today's most accurate 3D object detection methods make use of LIDAR sensor data \cite{Shi.20191231b,Yang.20190722,Shi.20191231} and surpass monocular object detection methods by a great margin on the KITTI data set \cite{Geiger.2012}. The leading lidar algorithm \cite{Shi.20191231b} achieves a 3D Average Precision (AP) of 81.43\,\% in the car category, whereas the leading monocular estimator \cite{Ding.20191213} achieves a 3D AP of 11.72\,\%. Regarding 2D object detection metrics, lidar and camera detection algorithms achieve comparable performance \cite{Geiger.2012}. The disadvantage of the camera sensors is the lack of 3D depth information in the 2D image representation. Nonetheless, object detection on a single camera sensor comes with numerous advantages, leading to broad research interest in the field in the recent years: In contrast to lidar sensors, the hardware availability of cameras for autonomous driving is greater due to lower sensor costs. The roads are designed for human vision which has a great comparability to camera data. The feature density in the camera data is greater than the one of the sparse lidar detections, which comes with a greater potential for the detection possibility. 

Furthermore, camera and lidar detection algorithms can be developed in a redundant manner, increasing the fault tolerance of the complete autonomous software stack. In the development process, a separated development for the different sensor modalities can lessen the overall complexity as the sensor specific development teams can work in their field of expertise independently. Early fusion approaches \cite{Ku.2018, Chen.2016, Nobis.2019}, while showing a great potential due to higher information density, come with the organizational drawback of requiring the knowledge of different sensor modalities at a low abstraction level in the whole team. 

\begin{figure}[ht]
	\includegraphics[width=8.6cm]{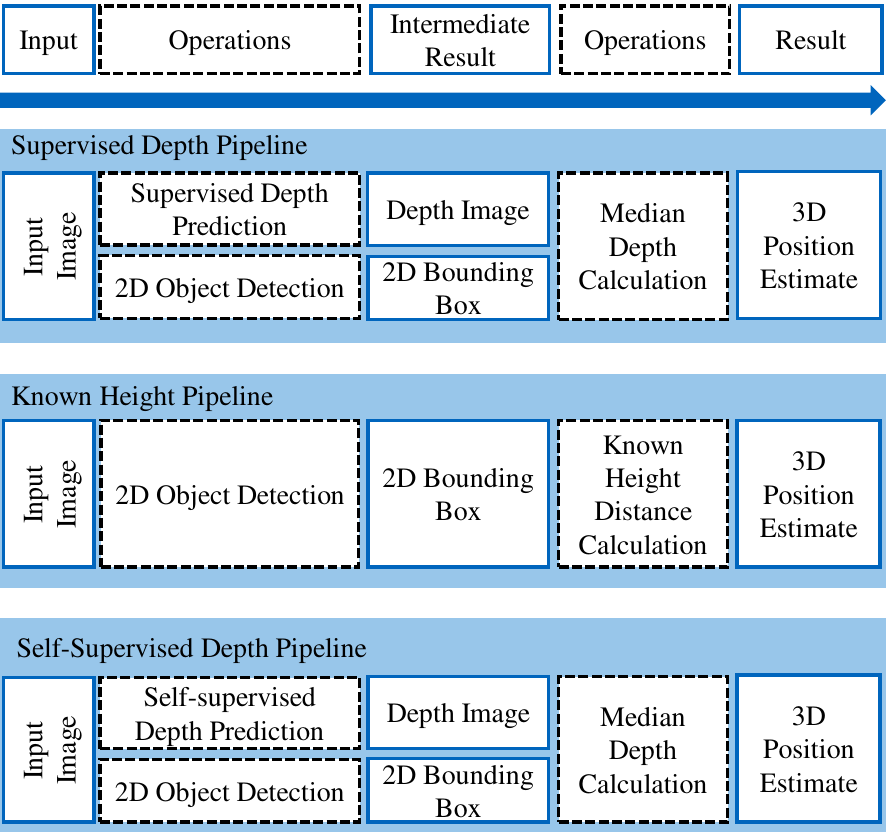}
\caption{3D object detection pipeline with three alternative configurations.}
\label{fig:pipeline}
\end{figure}

In this paper, we apply monocular 2D object detection and monocular depth estimation in a parallel pipeline to perform 3D object detection. Two stage 3D detection pipelines have been applied for example in \cite{Wang.20190614}. In comparison, we use a more simple approach, which estimates the depth of 2D bounding box detections and calculates the 3D position without estimating the full 3D pose. We compare three different depth estimation pipelines and evaluate their 3D position estimation performance on simulation and real world image data. An overview of the alternative detection pipeline configurations is given in Figure \ref{fig:pipeline}.

The detection performance is evaluated for the continuous trajectory of a race car on the track in simulation and in a real world scenario. The real world scenario was recorded in the context of the Roborace autonomous racing challenge \cite{Betz.2019, Roborace.2018}. 

Collecting appropriate data and labelling them correctly for camera learning algorithms can be tedious and error-prone. It would be beneficial if no extensive data collection and labelling strategy is necessary by using simulation data. Our work discusses the current drawbacks of bridging the domain gap between simulation and real world data. It hints to the potential to use simulation to adapt deep learning approaches for real world data sets. 

Section \ref{sec:related_work} discusses depth estimation, monocular object detection and simulation environments for autonomous driving. Section \ref{sec:development} presents the object detection pipeline developed in this work. Additionally, it gives insight to the data generation and data handling process with the simulator.
The evaluation and discussion of the approach is performed in Section~\ref{sec:results}. Finally, our conclusions from the work are presented in Section~\ref{sec:conclusions}.

\section{Related Work}
\label{sec:related_work}
In this section, we present the state of the art of: Monocular depth estimation,  2D monocular object detection and simulation environments for autonomous driving research.

\subsection{Monocular Depth Estimation}
Depth inference from monocular images is ill-posed. In recent years, different approaches emerged to deal with the lack of 3D information in images and to reconstruct the 3D scene \cite{Eigen.20140609, Ranftl.2016, Godard.2017, Mahjourian.2018, Casser.2019, Fu.2018, Alhashim.20190310}. The methods can be categorized as either supervised learning methods, which require ground-truth depth information, or self-supervised learning methods, which only require RGB images for training.

\cite{Godard.2017} reconstructs the monocular depth by learning the disparity for a virtual stereo setup from stereo image ground truth data. At inference time, only a monocular camera is necessary to reconstruct the 3D information. 

\cite{Mahjourian.2018} uses consecutive frames from a monocular camera to reconstruct adjacent frames through a neural network. The loss is constructed as an image reconstruction problem without an explicit depth term, thereby it does not require depth ground truth. To be able to reconstruct the frames, the network learns the transformation of the camera viewpoint explicitly and thereby also provides a source of odometry information. \cite{Casser.2019} augments their network to explicitly handle moving objects in the depth prediction. In the previous approaches, especially the depth estimation of objects moving at a similar speed as the camera resulted in those objects to be wrongly mapped to infinity. \cite{Gordon.20190410} extends this method in a way that it learns the intrinsic parameters of the source camera in addition to the depth estimate. 

The currently best performing network on the KITTI leaderboard is DORN \cite{Fu.2018} which uses an ordinal loss to calculate the depth for different discrete intervals. The rational behind this is to augment the influence of near depth values in the loss calculation, which are outweighted in the previous formulations by far depth values and increased depth estimation errors. 

DenseDepth \cite{Alhashim.20190310} present a loss function which takes the gradient of the depth into account for the loss calculation. This tackles the problem of edge-bleeding around the contours of objects. The shown performance metrics are slightly worse than the ones of DORN for the KITTI data set, while they surpass them for the NYU Depth v2 data set \cite{Silberman.2012}. The authors explain the worse performance on KITTI with the sparse ground truth information depth information in this data set. 

\subsection{Monocular 2D Object Detection}
Object detection on the 2D image space with deep learning method has seen a strong interest after early promising results of Overfeat \cite{Sermanet.20140224} and R-CNN \cite{Girshick.20141022}. An extensive review of 2D object detection methods is found in \cite{Zhao.20190416} and \cite{Huang.2016}.
 
\subsection{Simulation Environments}
A variety of simulation environments exists for the development of autonomous driving features \cite{Dosovitskiy.20171110,MATLAB.2018,UnityTechnologies.2019,Shah.20170718,IPGAutomotive.2019}. \cite{Rosique.2019} gives a further overview of perception systems and simulation environments. The use of simulation in the development of perception systems facilitates the data generation process. In simulation, a greater variety of scenes can be modeled. Edge cases \cite{Fridman.20180924} can be introduced explicitly into the data set. Additionally, the explicit modelling in the simulation enables the automatic generation of ground truth information. This is a great advantage to the time consuming or costly manual labelling for real world data sets \cite{Intel.2018, Wang.20190419}. While the usage of simulation and the benefits to real world perception are on a rise, current simulations still do not represent the real world environment in enough detail to make real world data collection and labeling obsolete.

\section{Object Detection Development}
\label{sec:development}
Depth estimation is the greatest challenge in 3D object detection with current methods. We implement three different strategies for the depth estimation of objects and analyze their advantages and drawbacks:
\begin{itemize}
\item Distance calculation using the 2D bounding box height, and the known height of the real world race car as a geometric constraint. We call this method known height assumption.
\item Depth estimation for the whole image using the supervised DenseDepth network. The distance to each object is calculated as the median depth value in the bounding box crop. Explicit knowledge about the objects, like height information, is not required in this approach.
\item Depth estimation for the whole image using the self-supervised struct2depth network. The distance to each object is calculated as the median depth value in the bounding box crop. Explicit knowledge about the objects, like height information, is not required in this approach.
\end{itemize}

To generate the 2D bounding box detection, we train and employ the one-stage network SSD \cite{Liu.2016} from the Tensorflow Object Detection API \cite{Huang.2016}. The performance of 2D object detection has been proven extensively in literature and is not the focus in this paper. For the evaluation results, we therefore mostly resort to ground truth 2D boxes to study the effect of the depth estimation isolated.

We perform 3D object detection by inferring the 3D position of the detected object by using different depth estimation strategies. In addition to the simple depth estimation we present in this paper, we employ 3D lidar detection networks \cite{Qi.20180413} on the same data in the underlying project, so that we are interested in keeping the similarities between the estimated camera depth data and lidar depth data high. For this, we are especially concerned about the edge bleeding problem from monocular depth estimation. Therefore, we adapt the DenseDepth network \cite{Alhashim.20190310} for our depth estimation pipeline. \cite{Uhrig.20170830} states that using synthetic data for training of depth estimation networks is an open challenge. To study the effect of synthetic training data more broadly, we implement a second depth estimation network in the pipeline. We integrate the struct2depth network \cite{Casser.2019}, as it is specialized to deal with object motion in the scene which occurs in the racing scenes of our use case. Furthermore, it is trained in a self-supervised manner, whereas DenseDepth requires depth ground truth information leading to a comparison of two inherently different estimation approaches. After using the depth networks, we calculate the distance to all objects detected in 2D. This is done by extracting the median depth of the 2D bounding box crop on the depth image. In the following, we describe the workings and main consideration to work with the different depth estimation approaches

\subsection{Known Height Assumption Pipeline}
The explicit distance calculation is possible since we are interested in calculating the distance to objects for which we know the real world height in meters $h_{car}$. Additionally we calculate the vertical focal length in pixel units $F_v$. The height of a 2D bounding box detection in pixel units is $H_{bb}$. The distance to the object in the bounding box can then be calculated using the following formula:

\begin{equation}
\label{eq:distance_calc_kh}
    d = h_{car} * F_v /H_{bb}.
\end{equation}

\subsection{Depth Network Pipelines}
The depth networks are trained following the training schemes from their original implementations and building on top of the publicly available network configurations trained on the KITTI data set.

\subsubsection{Self-supervised Pipeline}
The struct2depth network is trained on consecutive frames from racing scenes recorded in the simulator on two different race tracks. The camera vehicle follows the object vehicle in varying distances of up to \SI{100}{\metre}. In total there are around 4500 frames in the simulator training data set. The data generation is somewhat limited even in simulation due to the constraint that consecutive frames are needed for the training. On the other hand the network can be trained on simulator and real world data since no ground truth depth is necessary. Therefore in addition to the simulation data, the training data set contains around 4000 real world images.

\subsubsection{Supervised Pipeline}
The DenseDepth network is trained on simulator data on consecutive frames and additional arbitrary poses of the object vehicle. In the arbitrary scenes, the object is placed in a distance between \SIrange{4}{100}{\metre} in front of the vehicle with arbitrary rotations between \SIrange{-90}{+90}{\degree} relative to the ego vehicle. The poses are recorded for numerous locations around the race track to generate a great variety in the data set.

\subsection{Computational Considerations}
The simple distance calculation is computationally negligible. Whereas the depth estimation networks need additional GPU resources. For practical considerations, they can be run in parallel to the 2D object detection networks, so that the overall delay for a real time inference is the maximum of the 2D detection network and the depth estimation network inference time and not the addition of the two.

\subsection{Simulation Design}

A suitable data set must contain realistic environment conditions and sufficiently large variety in order to generalize well in the simulation and real world domain. The evaluation on the real world data is performed on a stretch of a race track which consists of a left-right curve combination and a straight. The simulation models the same race track and an additional race track from the Roborace competition from GPS locations of the real race track bounds. Additionally, videos of the race track are analyzed to create a more realistic environment replication by including trees, hills and buildings which are present along the race track. Different lighting conditions are simulated. This includes lens flare and different positioning of the sun and sky modelling. The paintwork of the race car is varied in three different setups to make the network invariant to the specific paint of the vehicle. As simulation backbone, we use the Unity environment, because of the ease-of-use of its functionalities and the appealing graphics performance.

\subsection{Data sets}

Both networks are trained on a data set generated in our Unity race simulation. The simulation environment is programmed to output the ground truth 3D poses of the ego and the object vehicles for every frame. The simulation saves images captured by a virtual camera which is configured to match the intrinsic and extrinsic parameters of the real world data set. Furthermore, it delivers the pixel-level ground truth depth and segmentation mask for the object vehicles. 

The ground truth depth is saved in a 16bit PNG format. The resolution of a standard 8bit format is not fine-grained enough to store the depth information for the range of interest up to \SI{100}{\metre}. The PNG format uses a lossless compression to prevent depth artifacts which occur in JPG images. 

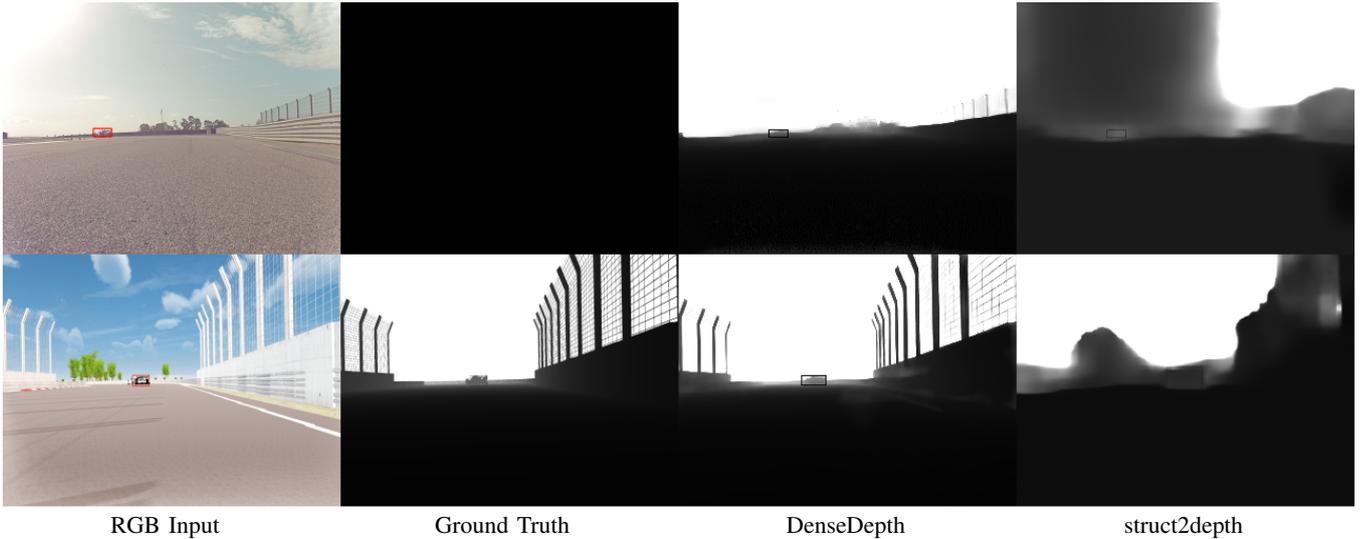
\begin{figure*}[ht]
    \centering
    \begin{overpic}[width=\textwidth]{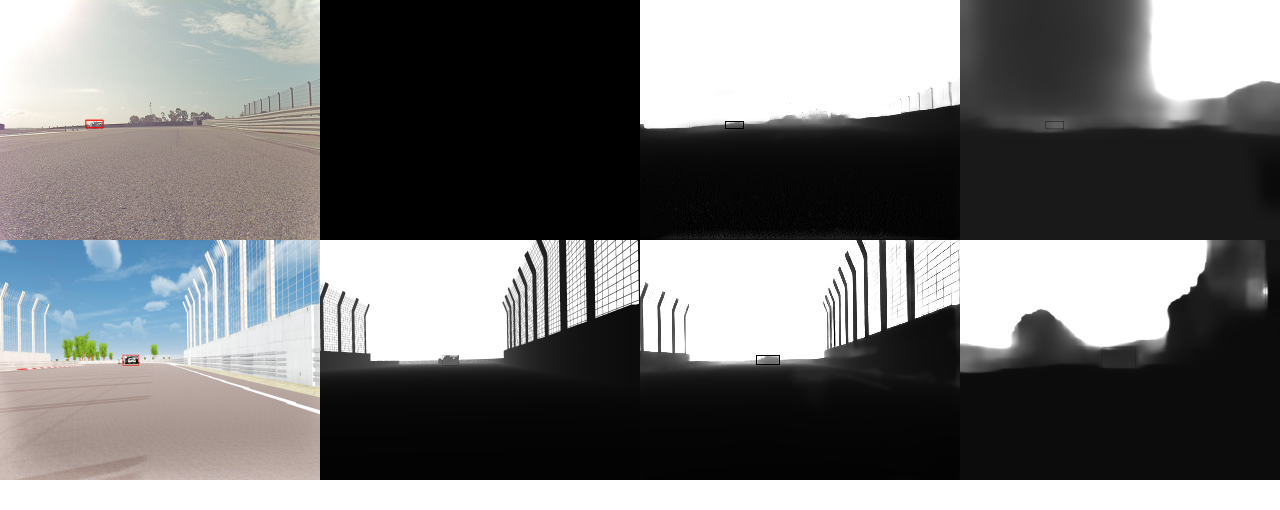}
     \put (8,0.3) {\small RGB Input}
     \put (32,0.3) {\small Ground Truth}
     \put (58,0.3) {\small DenseDepth}
     \put (83,0.3) {\small struct2depth}
    \end{overpic}
    \caption{Example scenes and corresponding depth images. The first row shows real world data. The second row shows simulation data.}
    \label{fig:data_overview}
\end{figure*}

In the real world recording the original camera resolution is scaled down by a factor of two to enable real-time recording of the image data on the vehicle hardware. As a difference, in the simulation the full resolution of the modeled camera is used for recording and known height inference. The depth networks operate with the same downscaled resolution for simulation and real world data.

The simulation data set contain images from two different race tracks. In the simulation training sets, the camera vehicle follows an object vehicle at arbitrary distances of up to \SI{100}{\metre}. Furthermore, the self-supervised depth network is trained with real world race track data.

\section{Experiments and Results}
\label{sec:results}

This section evaluates the results of the three different depth estimation techniques in detail on three different test sets.

\subsection{Evaluation Metrics}

The networks are evaluated regarding the 3D recall and Average Translational Error (ATE) metrics. These 3D metrics are inspired by the 3D mAP proposed by the authors of \cite{Caesar.2019}. The 3D recall is the main result of the evaluation. The use of an additional precision or AP metric is not relevant here, since 2D false positives do not occur on the whole data set. All 3D false positive (FP) detections arise from errors in the distance estimation of true positives (TP) which are therefore already registered in the 3D recall metric. The ATE gives an additional insight into the absolute translational error in the ground plane for both the TP and the FP detections.

\subsection{Evaluation}

An exemplary frame from the real world test set is shown in the first row in Figure \ref{fig:data_overview}. Row two shows the same scene modeled in the simulation. 

The test of the network is performed in three different environments: 
\begin{itemize}
    \item Test set 1 : Vehicle following sequence at distances of around \SI{20}{\metre} in the simulation. 355 frames.
    \item Test set 2 : Vehicle following sequence at distances of around \SI{70}{\metre} in the simulation. 209 frames.
    \item Test set 3 : Vehicle following sequence at distances of around \SI{50}{\metre} in the real world scenario. 738 frames.
\end{itemize}

\begin{table}[ht]
\caption{3D position detection with different pipeline configurations. If not stated otherwise, the 2D bounding box is taken from the ground truth boxes.
GT: Ground truth depth. DD: DenseDepth. S2D: struct2depth. KH: Known height assumption. SSD: 2D bunding box detection with single shot detector.}
\label{table:depth_comparision}
\centering
\begin{tabular}{|c | c | c| c |} 
 \hline
 Test set & Depth  & 3D Recall & ATE \\ [0.5ex] 
  & Configuration &  in \% & in m \\ [0.5ex] 
 \hline\hline
 1 - Sim 20m & GT & 95.99 & 0.2\\
         & KH & 72.32 & 0.8  \\
         & S2D & 46.34 & 2.2  \\ 
         & DD & 32.25 & 4.4\\ 

 \hline
  2 - Sim 70m   & GT & 90.55 & 0.8\\
             & KH & 55.98 & 1.5  \\
             & S2D & 14.95 & 9.3  \\ 
             & DD & 7.54 & 11.4\\ 

  \hline
3 - Real World & KH  & 17.68 & 4.9  \\
           & S2D & 2.57  & 23.7  \\
           & DD  & 2.37  & 15.6  \\ 
           & SSD KH & 27.71 & 3.6 \\
           \hline
\end{tabular}
\end{table}

For each scene, we evaluate the object detection pipeline with different configurations for the distance estimation. The results are shown in Table \ref{table:depth_comparision}.

\begin{figure}[ht]
	\input{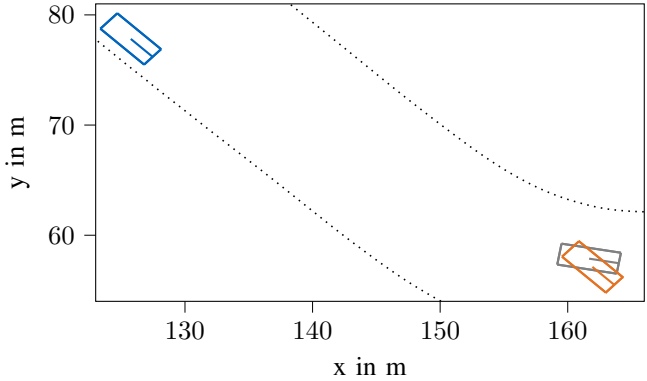}
\caption{The ego vehicle position is shown in blue. The detection object position in orange is calculated with ground truth 2D bounding box and ground truth depth information. Ground truth object position in gray. Due to the unknown heading of the detected vehicle, the center point detection has a slight error in the curve.}
\label{fig:gt_error}
\end{figure}

For the simulator data, we first calculate the 3D position estimation by using the ground truth 2D bounding boxes and calculating the median distance for this bounding box crop from the ground truth depth image. This is the theoretical maximum 3D recall that we can achieve with the presented 2D + depth approaches if we would be able to generate perfect depth from the image data. We do not achieve a perfect 3D recall since we compare the center of the opponent bounding box to the ground truth. Since all methods measure the distance to the back of the objects, we have to add an offset to represent the center point of the object vehicle. This offset is known if we know the vehicle size and assume a straight heading of the vehicle in front of us. In curves however, this assumption introduces an error due to the unknown heading of the object vehicle. We do not introduce a tracking of the position and heading to mitigate this error, since we want to evaluate the raw detection performance without the augmentation of the results through tracking. The error by the center offset in a curve is shown in Figure \ref{fig:gt_error}.

Table \ref{table:depth_comparision} shows a general decline in accuracy for an increase of the distance to the object and to the detection on real world data.

\subsubsection{Known Height Assumption Discussion}

The known height distance calculation achieves the best results over all data sets. Even though the use case of this form of distance calculation is limited and the deep learning approaches provide a lot of more information for the whole scene, it fits best for the shown use case and the available data. As a drawback of the method due to the discrete pixel size, the detection resolution declines with an increased distance of the object vehicle and with lower camera resolutions. With a camera of focal length \SI{900}{px} we can detect an object of \SI{1}{\metre} height at a distance of \SI{50}{\metre}. The next farther detectable distance bin for the same object is at almost \SI{53}{\metre}. As seen in Equation \ref{eq:distance_calc_kh}, the distance estimation scales inversely proportional to the measured pixel height. At a distance of \SI{25}{\metre}, the next farther distance bin lies at around \SI{25.7}{\metre}. Increasing image resolutions and thereby increasing the focal length in pixel units, increases the distance resolution. At the same time, this leads to a greater amount of raw data to process and more expensive cameras. 

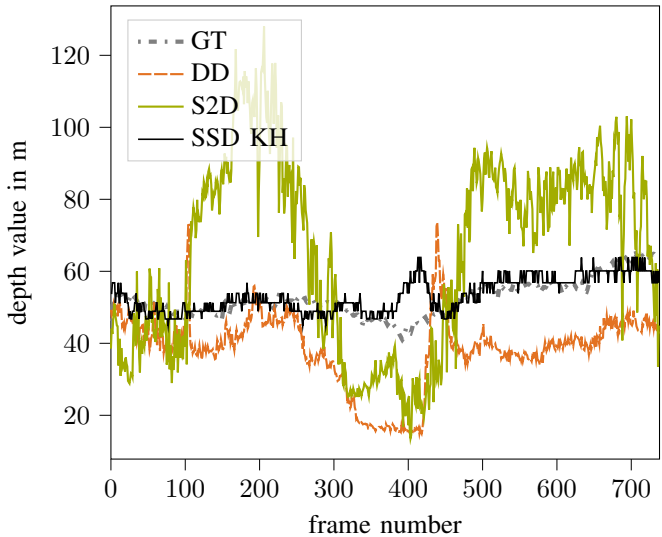
\begin{figure}[ht]
	\input{figures/depth_rw.tex}
\caption{Depth estimation results for real world test set 3.}
\label{fig:dd_rw}
\end{figure}

The resolution in our real world data set is half of the full HD solutions used in the simulator. The errors for this method are thereby increased for the real world data set. The discrete distance bins for the known height distance calculation can be observed in Figure \ref{fig:dd_rw}. A single pixel error in height estimation, can already lead to a FP 3D detection. Additionally, the 2D bounding boxes are annotated manually and therefore the ground truth bounding box height in pixels is not always accurate as it is in the simulation. It is interesting to note, that the 3D recall metric for the 2D bounding box generated by the SSD detection network surpasses the one of the ground truth bounding box. Concerning the height measurement, the 2D detector seems to achieve a better performance than our manual labelling. A third source of error for this method is introduced due to relative non-zero pitch angles of the detected vehicles. The 2D bounding box will naturally have a greater height if it has to encompass an inclined vehicle. Even if the 2D detection works flawless, the distance will thereby be underestimated, e.g. for a vehicle driving up a hill.

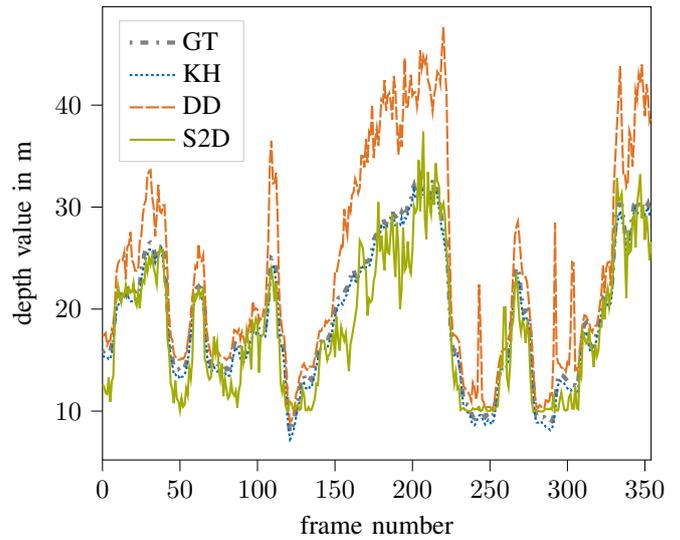
\begin{figure}[ht]
	\input{figures/depth_20m.tex}
\caption{Depth estimation results for simulation test set 1.}
\label{fig:dd_20m}
\end{figure}

\begin{figure}[ht]
	\input{figures/depth_70m.tex}
\caption{Depth estimation results for simulation test set 2.}
\label{fig:dd_70m}
\end{figure}
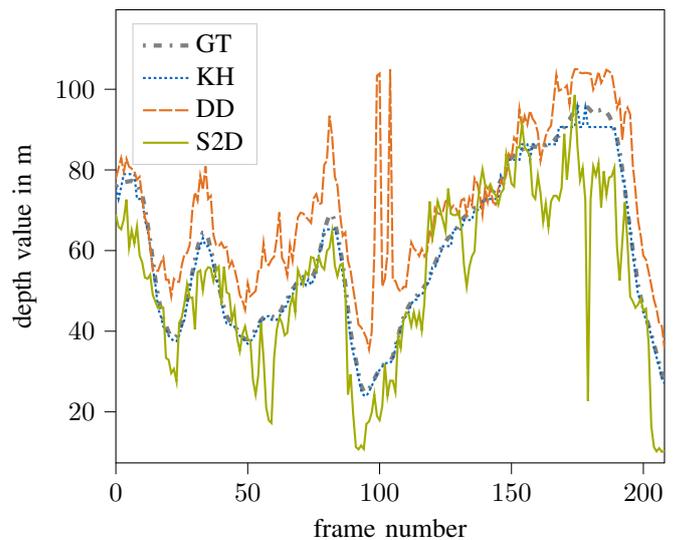

\subsubsection{Supervised Depth Discussion}

The DenseDepth estimation leads to the worst recall results. However, an optical assessment of the depth estimation images show a reasonable scene representation. The general distance estimation feasibility can be seen in Figures \ref{fig:dd_20m} and \ref{fig:dd_70m} for the simulator test sets. Even though the estimation is not accurate enough to achieve an accurate 3D recall metric, the general distance estimation trend follows the ground truth value for the detected object.

The distance outliers around frame 100 in Figure \ref{fig:dd_70m} result from images where the object vehicle is barely visible at the edge of the frame, making the depth estimation as well as the median calculation prone to greater errors. 

The supervised depth pipeline overestimates the depth for most frames.

The distance error peaks in the sequence shown in Figure \ref{fig:dd_20m} at short distance the error peaks, are found due to artifacts in the estimated depth map which are most notably present between frames 150 to 230. The reason for these artifacts is unknown.

The appliance to the real world test data does not lead to comparable results. This is shown in Figure \ref{fig:dd_rw}. The distance estimate is varying greatly from the ground truth. An optical inspection leaves a general consistent impression of the environment. However, the race car is less recognizable than it is for the simulation images. During the frames around 230 to 400, the vehicle is driving towards the sun. This coincides with the drop in depth detection performance. Even though similar lighting effects are modelled in the simulation, it seems the network can not generalize well to these conditions.

Additionally the object vehicle is always partly over the horizon line in the camera frame of the simulation. Due to the hilly terrain this is rarely the case in the real world sequence. An accurate terrain modelling in the simulation training data could potentially lead to better results here. The optical impression from the DenseDepth output shown in Figure \ref{fig:data_overview} motivates the conclusion, that the network is generally able to learn the depth of its environment from simulation even when it is applied to real world data. However, a high variety needs to be covered in the training data set. In line with comparable depth networks, this network learns a strong prior about its environment. This comes with the drawback of weak generalizability of the results to additional data sets or unknown scenarios.

\subsubsection{Self-supervised Depth Discussion}

The struct2depth inference results in the second best recall metrics. The optical assessment of the depth images, show that the vehicle is distinguishable from the environment, however the edges are not visible clearly. Additionally, the optical impression of the environment estimation shown in Figure \ref{fig:data_overview} is worse than the one of the DenseDepth. The visual appealing results from the paper could not be reproduced. The visualizations shown in the original paper seem to performs well for the close range of roughly \SI{20}{\metre} from the vehicle, farther distances seem to be learned worse by this approach. 

The detection performance in the real world data set is not promising as shown in Figure \ref{fig:dd_rw}. Even though it is trained on images from simulation and on real world data, it can not generalize to the real world data set on testing. While the metric results are comparable, the optical impression of the scene is not as well represented as it is in the supervised approach. Similar to DenseDepth, the generalization of the network seems to be weak between the different data sets: KITTI, Unity simulation and the real world race track scenario.


\section{Conclusions}
\label{sec:conclusions}
This paper investigates the capabilities of monocular camera systems for 3D object detection. Firstly, a simulation environment with the Unity 3D engine is built to simulate autonomous driving scenarios on race track environments. This simulator generates image and ground truth data to train neural networks for depth estimation. An algorithmic distance calculation, a self-supervised deep learning method and supervised deep learning method are implemented to showcase limits and possibilities of 3D monocular object detection. 

The depth network results can be generalized over data sets to a limited extent, e.g. general optical impression. If the camera intrinsics or resolution of the input images are changed, the level of generalizability is further reduced. The learning works only within specific conditions. Extrapolating the results to a greater amount of training data, we conclude that 3D object detection could be performed reasonable well with current methods for close range scenarios of around \SI{20}{\metre} distances. The detection performance deteriorates for greater distances. In theory this can be compensated by higher image resolutions at the cost of a higher overall data rate. Alternatively cameras with a small and wide field-of-views can be combined to enable accurate detections for additional desired ranges. Leading to higher system complexity and package requirements.

For future work, the level of detail of the simulation needs to be augmented. Especially modelling the height of the race track terrain is expected to make the depth networks perform better in real world scenarios, since the system has not seen vertical movement of the camera in the training. Even though the known height calculation showed the best performance for the application, it is not easily applicable to more general scenarios with a number of different unknown objects. In the current study, the self-supervised network showed the better overall performance in terms of metrics. The supervised approach produced a visually more consistent depth map even though the self-supervised approach was additionally trained on a real world data. The detection results after training with additional simulator details and a more diverse data set need to be investigated for both configurations.

\cite{Alhaija.20170804} augments the KITTI data set scenes with additional object instances to generate more diverse training data. The augmentation of real world data with simulation objects is another promising approach to bridge the domain gap.

Even though there is a lot of research conducted in this direction, 3D monocular object detection is not on par with the performance of stereo camera or lidar methods even in favorable conditions for the monocular camera. The networks learn a strong prior about the environment and can create realistic 3D models for a specific environment. However, after an extensive literature research there seems to be no current model that is able to generalize well in a variety of conditions. \cite{Chang.20190418} states that the accuracy of depth estimation is heavily data set depend. A real world series application of a monocular depth estimation approach therefore would need to incorporate vast amounts of training data, covering all possible future scenarios yet in the training. While this could be possible with fleet data recordings, it would still be a tedious task. With current methods, the fusion of camera information with distance measuring sensors such as lidar and radar still seems to be the most effective method to perform object detection in 3D.


\section*{Contributions and Acknowledgement}
Felix Nobis initiated the idea of this paper and contributed essentially to its conception and content. Fabian Brunhuber contributed to the development of the simulation environment and the training of the self-supervised depth network. Simon Janssen contributed to the implementation of the inference pipeline and the training of the supervised depth network. Johannes Betz revised the paper critically. Markus Lienkamp made an essential contribution to the conception of the research project. He revised the paper critically for important intellectual content. He gave final approval of the version to be published and agrees to all aspects of the work. As a guarantor, he accepts the responsibility for the overall integrity of the paper. We express gratitude to Continental Engineering Service for funding for the underlying research project.


\bibliographystyle{./bibliography/IEEEtran} 
\bibliography{./bibliography/bibliography}

\addtolength{\textheight}{-12cm}   

\end{document}

%% file: figures/depth_20m.tex
\begin{tikzpicture}

\definecolor{color0}{rgb}{0,0.396078431372549,0.741176470588235}
\definecolor{color1}{rgb}{0.890196078431372,0.447058823529412,0.133333333333333}
\definecolor{color2}{rgb}{0.635294117647059,0.67843137254902,0}

\begin{axis}[
legend cell align={left},
legend style={fill opacity=0.8, draw opacity=1, text opacity=1, at={(0.03,0.97)}, anchor=north west, draw=white!80!black},
tick align=outside,
tick pos=left,
width=8.8cm,
x grid style={white!69.0196078431373!black},
xlabel={frame number},
xmin=0, xmax=354,
xtick style={color=black},
y grid style={white!69.0196078431373!black},
ylabel={depth value in m},
ymin=5.2087, ymax=49.6333,
ytick style={color=black}
]
\addplot [ultra thick, white!50.1960784313725!black, dash pattern=on 1pt off 3pt on 3pt off 3pt]
table {%
0 15.93717
1 15.87062
2 15.98965
3 15.9387
4 15.72535
5 15.89103
6 16.61121
7 17.60925
8 19.23753
9 20.69402
10 21.36764
11 21.36678
12 21.77888
13 21.64868
14 21.88516
15 21.55222
16 22.30018
17 21.45143
18 21.43092
19 21.33375
20 21.3157
21 21.69887
22 21.79568
23 21.77205
24 22.27257
25 23.74891
26 24.34271
27 25.40009
28 25.91294
29 26.07506
30 26.3947
31 26.52524
32 26.40045
33 25.90056
34 25.27806
35 25.65646
36 26.13634
37 26.10759
38 26.189
39 26.30129
40 24.62481
41 22.57544
42 19.82261
43 17.80108
44 15.98109
45 15.18814
46 14.34085
47 14.31675
48 14.00851
49 14.08628
50 13.85061
51 14.03725
52 14.03682
53 14.27681
54 14.69847
55 15.48351
56 16.85237
57 18.38705
58 20.03536
59 21.36834
60 21.8803
61 22.09618
62 22.24488
63 22.08822
64 22.02104
65 21.55389
66 19.92919
67 17.34196
68 15.54938
69 15.05507
70 14.55223
71 14.65842
72 14.99545
73 14.71879
74 14.596
75 14.62701
76 14.47818
77 14.4571
78 14.2361
79 14.39141
80 14.08408
81 14.11167
82 14.21968
83 14.66749
84 15.75574
85 16.53368
86 16.95348
87 16.57653
88 16.55842
89 15.82952
90 15.70514
91 15.38421
92 15.65887
93 15.86871
94 16.36571
95 17.25897
96 17.36109
97 18.25848
98 18.45205
99 18.27197
100 18.44003
101 17.88161
102 17.97843
103 18.03946
104 17.94927
105 18.15901
106 19.23888
107 21.03278
108 23.44909
109 25.00374
110 24.89295
111 24.64417
112 23.78139
113 21.78856
114 18.79381
115 17.16553
116 16.61034
117 16.03694
118 14.29447
119 11.36818
120 8.958755
121 7.875684
122 8.205788
123 8.712335
124 9.294602
125 10.24946
126 10.99883
127 11.52742
128 12.68878
129 13.33691
130 13.47451
131 12.99658
132 12.9456
133 12.81535
134 13.12999
135 13.0178
136 13.60156
137 14.3422
138 14.94945
139 15.64911
140 16.5579
141 16.80054
142 16.79701
143 16.71407
144 16.58095
145 16.70845
146 17.27298
147 17.60182
148 18.27193
149 19.213
150 19.76612
151 20.60864
152 21.00415
153 21.14543
154 21.15828
155 21.16357
156 21.58693
157 21.94979
158 22.16915
159 22.70326
160 23.05191
161 23.85708
162 23.84023
163 23.31241
164 23.16605
165 23.69952
166 24.00279
167 24.37623
168 24.74499
169 24.57616
170 24.67369
171 24.64214
172 24.74957
173 25.33853
174 25.78445
175 26.80365
176 27.11936
177 27.35852
178 28.09623
179 28.09383
180 28.87077
181 28.86115
182 28.44882
183 28.33908
184 28.83401
185 28.65189
186 28.71819
187 29.68121
188 29.17643
189 29.10478
190 28.854
191 28.83402
192 29.02352
193 29.44008
194 29.01459
195 29.47682
196 29.49959
197 30.10625
198 30.48956
199 30.66135
200 31.67093
201 32.08182
202 31.32734
203 31.29717
204 32.3366
205 33.09638
206 32.67161
207 31.84788
208 31.70633
209 31.79774
210 31.56685
211 31.09871
212 31.54892
213 32.46512
214 32.46327
215 32.41721
216 32.43973
217 30.88741
218 29.69381
219 28.67452
220 27.92556
221 27.38955
222 26.22979
223 23.73307
224 20.20487
225 17.53985
226 16.17028
227 15.9521
228 16.33185
229 16.02433
230 15.38919
231 14.17508
232 12.47522
233 11.01337
234 10.71364
235 10.65747
236 10.56238
237 10.39948
238 9.733485
239 9.4285
240 9.29614
241 9.510444
242 10.04316
243 10.27736
244 10.31475
245 10.01743
246 9.620374
247 9.420552
248 9.615105
249 9.694519
250 9.602854
251 9.295162
252 9.572043
253 10.22402
254 11.2706
255 12.74593
256 13.71236
257 14.65678
258 14.80215
259 14.86849
260 14.97282
261 15.16367
262 15.79058
263 17.13724
264 19.39309
265 21.23583
266 23.36351
267 24.09212
268 24.084
269 23.33002
270 21.54032
271 20.53448
272 19.70564
273 19.79813
274 19.87846
275 19.36551
276 17.83704
277 15.20711
278 12.49968
279 10.27621
280 9.477324
281 9.46483
282 9.522781
283 9.651062
284 9.401614
285 9.402747
286 9.193235
287 9.119478
288 8.973829
289 8.819701
290 9.018396
291 9.441764
292 10.93065
293 12.38971
294 13.26902
295 13.67713
296 13.51773
297 13.47321
298 13.3412
299 13.14477
300 12.96972
301 12.80115
302 12.83351
303 12.58521
304 12.55174
305 12.54165
306 12.35854
307 12.96808
308 14.48418
309 16.62043
310 17.93521
311 18.50168
312 18.65421
313 17.7627
314 16.98999
315 16.41849
316 16.30719
317 16.50844
318 17.02214
319 17.8133
320 18.46794
321 19.25109
322 20.10332
323 21.00127
324 20.79795
325 21.08035
326 21.28766
327 20.98968
328 21.57072
329 22.81067
330 25.54263
331 28.02206
332 29.71316
333 30.26605
334 30.29194
335 30.20054
336 28.99825
337 27.72517
338 26.43499
339 26.34356
340 27.12228
341 28.14073
342 29.27943
343 30.23802
344 30.37576
345 29.94564
346 30.24244
347 30.18299
348 30.10377
349 30.06396
350 29.94558
351 30.03441
352 30.35232
353 29.76312
354 29.79189
};
\addlegendentry{GT}
\addplot [thick, color0, dash pattern=on 1pt off 1pt]
table {%
0 15.401
1 15.335
2 15.335
3 15.335
4 15.077
5 15.077
6 15.747
7 16.936
8 18.645
9 19.9
10 20.323
11 20.77
12 21.083
13 21.083
14 21.244
15 21.083
16 21.748
17 20.925
18 20.77
19 20.77
20 20.77
21 21.083
22 21.244
23 20.925
24 21.244
25 22.854
26 23.676
27 24.345
28 25.062
29 25.313
30 25.835
31 25.835
32 25.835
33 25.313
34 24.578
35 25.062
36 25.571
37 25.571
38 25.835
39 24.817
40 23.256
41 21.922
42 18.999
43 16.936
44 15.204
45 14.599
46 13.579
47 13.625
48 13.314
49 13.401
50 13.147
51 13.357
52 13.401
53 13.625
54 14.01
55 14.832
56 16.193
57 17.684
58 19.245
59 20.469
60 21.083
61 21.576
62 21.576
63 21.576
64 21.244
65 20.469
66 18.879
67 16.43
68 14.599
69 14.268
70 13.911
71 14.01
72 14.322
73 14.01
74 13.862
75 13.911
76 13.766
77 13.718
78 13.534
79 13.718
80 13.401
81 13.445
82 13.489
83 13.96
84 15.077
85 15.819
86 16.271
87 15.891
88 15.819
89 15.14
90 14.953
91 14.773
92 15.077
93 15.335
94 15.747
95 16.677
96 16.762
97 17.684
98 17.784
99 17.585
100 17.885
101 17.298
102 17.392
103 17.392
104 17.298
105 17.298
106 18.419
107 20.323
108 22.469
109 24.117
110 24.345
111 23.676
112 22.66
113 20.618
114 17.784
115 16.116
116 16.04
117 15.204
118 13.445
119 10.658
120 8.269
121 7.228
122 7.527
123 8.01
124 8.568
125 9.527
126 10.269
127 10.822
128 11.983
129 12.609
130 12.755
131 12.266
132 12.201
133 12.106
134 12.468
135 12.4
136 12.946
137 13.766
138 14.431
139 15.077
140 15.965
141 16.193
142 16.193
143 16.193
144 16.04
145 16.116
146 16.677
147 17.024
148 17.684
149 18.645
150 19.245
151 19.764
152 20.323
153 20.618
154 20.618
155 20.469
156 20.925
157 21.244
158 21.576
159 22.101
160 22.469
161 23.053
162 23.256
163 22.854
164 22.469
165 23.053
166 23.464
167 23.894
168 24.117
169 23.894
170 24.117
171 24.117
172 24.117
173 24.578
174 25.313
175 26.106
176 26.67
177 26.67
178 27.264
179 27.264
180 28.22
181 28.22
182 27.893
183 27.893
184 28.22
185 28.22
186 28.22
187 29.261
188 28.904
189 28.22
190 28.22
191 28.22
192 28.22
193 28.904
194 28.22
195 28.904
196 28.904
197 29.261
198 30.009
199 30.009
200 30.804
201 31.651
202 30.804
203 30.804
204 32.095
205 32.555
206 32.095
207 31.22
208 31.22
209 31.22
210 30.804
211 30.804
212 30.804
213 32.095
214 32.095
215 31.651
216 31.22
217 30.009
218 28.904
219 27.574
220 27.264
221 26.106
222 25.313
223 23.053
224 19.5
225 16.677
226 15.269
227 15.401
228 15.676
229 15.468
230 14.656
231 13.579
232 11.894
233 10.321
234 10.07
235 9.96
236 9.841
237 9.672
238 9.006
239 8.735
240 8.617
241 8.815
242 9.331
243 9.565
244 9.605
245 9.319
246 8.908
247 8.683
248 8.937
249 9.152
250 8.986
251 8.692
252 8.986
253 9.605
254 10.699
255 12.169
256 13.106
257 14.01
258 14.163
259 14.268
260 14.268
261 14.486
262 14.953
263 16.43
264 18.645
265 20.618
266 22.283
267 23.053
268 23.053
269 22.469
270 20.77
271 19.631
272 18.999
273 19.245
274 18.999
275 18.419
276 16.762
277 14.599
278 11.749
279 9.539
280 8.889
281 8.861
282 8.899
283 9.006
284 8.726
285 8.683
286 8.46
287 8.401
288 8.269
289 8.128
290 8.316
291 8.726
292 10.218
293 11.666
294 12.538
295 12.985
296 12.907
297 12.907
298 12.681
299 12.434
300 12.234
301 12.075
302 12.138
303 11.864
304 11.835
305 11.806
306 11.638
307 12.234
308 13.718
309 15.891
310 17.206
311 17.784
312 17.684
313 17.206
314 16.271
315 15.747
316 15.747
317 15.891
318 16.43
319 17.298
320 17.784
321 18.645
322 19.5
323 20.323
324 20.179
325 20.323
326 20.618
327 20.179
328 20.618
329 21.922
330 25.062
331 26.963
332 28.557
333 28.904
334 29.629
335 28.904
336 28.22
337 27.264
338 25.062
339 25.313
340 26.106
341 27.574
342 28.22
343 29.629
344 30.009
345 29.261
346 29.629
347 29.629
348 29.629
349 29.629
350 29.261
351 29.629
352 30.009
353 29.261
354 29.261
};
\addlegendentry{KH}
\addplot [thick, color1, dash pattern=on 5pt off 1pt]
table {%
0 17.353
1 17.45
2 17.634
3 16.49
4 16.863
5 16.889
6 18.005
7 17.753
8 20.587
9 23.607
10 24.477
11 24.763
12 24.652
13 24.804
14 25.524
15 22.91
16 26.12
17 26.302
18 27.544
19 24.786
20 24.501
21 23.76
22 23.737
23 23.779
24 26.621
25 27.144
26 29.158
27 30.279
28 30.718
29 33.122
30 33.509
31 33.617
32 31.094
33 29.977
34 27.152
35 29.187
36 32.197
37 29.587
38 29.42
39 30.176
40 30.206
41 27.058
42 23.291
43 19.281
44 17.261
45 17.037
46 15.632
47 15.311
48 15.282
49 15.038
50 15.035
51 15.173
52 15.12
53 15.32
54 15.843
55 16.732
56 17.869
57 18.96
58 21.098
59 22.677
60 24.252
61 24.093
62 26.384
63 24.791
64 24.673
65 25.266
66 22.166
67 18.106
68 16.916
69 16.156
70 15.421
71 15.472
72 15.766
73 15.606
74 15.85
75 15.687
76 15.4
77 15.332
78 15.295
79 15.174
80 15.025
81 15.21
82 15.293
83 15.856
84 16.642
85 17.95
86 17.877
87 17.646
88 17.975
89 17.06
90 17.728
91 17.144
92 16.604
93 17.142
94 17.753
95 18.84
96 18.284
97 20.74
98 19.931
99 20.073
100 19.954
101 18.754
102 19.089
103 18.997
104 19.895
105 19.698
106 23.122
107 26.912
108 34.451
109 36.485
110 32.322
111 32.406
112 33.336
113 29.02
114 21.322
115 19.183
116 17.842
117 17.537
118 15.806
119 12.635
120 9.883
121 8.917
122 9.096
123 9.56
124 10.364
125 11.73
126 12.125
127 13.289
128 13.655
129 14.411
130 14.64
131 14.222
132 14.057
133 14.407
134 14.363
135 14.334
136 14.413
137 15.28
138 16.091
139 17.023
140 17.85
141 17.839
142 18.13
143 18.112
144 18.842
145 18.672
146 18.918
147 18.946
148 19.328
149 21.008
150 23.659
151 23.586
152 24.305
153 25.364
154 26.159
155 25.875
156 29.791
157 28.152
158 27.267
159 30.164
160 29.586
161 31.193
162 31.445
163 32.747
164 32.193
165 31.363
166 31.903
167 35.025
168 34.984
169 33.904
170 36.693
171 34.352
172 37.318
173 36.9
174 40.052
175 34.814
176 36.594
177 37.692
178 36.004
179 38.133
180 40.53
181 40.304
182 42.436
183 39.444
184 40.888
185 41.177
186 40.262
187 38.598
188 42.81
189 41.5
190 37.938
191 35.16
192 38.186
193 35.895
194 39.584
195 44.574
196 39.813
197 39.863
198 41.508
199 42.971
200 40.951
201 41.265
202 41.245
203 39.871
204 41.185
205 45.354
206 43.634
207 44.757
208 44.122
209 44.58
210 42.416
211 41.927
212 40.836
213 39.415
214 40.966
215 41.847
216 43.265
217 43.032
218 42.169
219 45.383
220 47.614
221 43.679
222 41.132
223 32.255
224 25.948
225 20.049
226 17.626
227 17.451
228 17.362
229 17.121
230 16.842
231 15.174
232 13.817
233 12.403
234 12.264
235 11.982
236 11.827
237 11.488
238 10.974
239 12.793
240 12.426
241 11.101
242 11.959
243 22.398
244 17.881
245 11.136
246 10.848
247 10.35
248 10.377
249 10.401
250 10.402
251 10.255
252 10.277
253 11.134
254 12.416
255 13.703
256 14.818
257 15.598
258 15.927
259 16.059
260 16.044
261 16.484
262 17.217
263 17.633
264 20.417
265 23.561
266 27.005
267 28.032
268 28.559
269 27
270 24.931
271 23.457
272 23.118
273 23.156
274 22.551
275 20.895
276 18.915
277 16.643
278 14.02
279 11.518
280 10.451
281 10.3
282 10.588
283 10.449
284 10.243
285 10.198
286 11.064
287 11.025
288 10.742
289 11.702
290 11.571
291 15.342
292 28.485
293 15.365
294 14.709
295 14.599
296 14.281
297 14.215
298 14.299
299 13.9
300 14.751
301 14.399
302 17.422
303 24.607
304 24.505
305 14.04
306 14.194
307 13.8
308 15.921
309 17.42
310 18.7
311 19.402
312 19.332
313 18.955
314 18.773
315 17.944
316 17.741
317 18.262
318 18.354
319 18.389
320 19.297
321 20.485
322 21.664
323 24.324
324 21.602
325 23.95
326 24.571
327 24.624
328 22.333
329 26.105
330 32.276
331 35.101
332 38.125
333 40.685
334 43.833
335 38.681
336 33.332
337 33.412
338 32.002
339 32.445
340 37.403
341 36.715
342 33.978
343 36.668
344 41.398
345 40.573
346 42.974
347 42.115
348 43.971
349 41.226
350 38.924
351 42.044
352 40.935
353 38.409
354 38.909
};
\addlegendentry{DD}
\addplot [thick, color2]
table {%
0 12.602
1 12.424
2 11.793
3 11.662
4 13.293
5 10.921
6 12.901
7 13.457
8 20.29
9 20.869
10 22.279
11 21.289
12 20.633
13 21.478
14 21.254
15 22.364
16 21.52
17 22.057
18 21.606
19 21.704
20 22.549
21 20.87
22 20.35
23 21.626
24 20.077
25 21.892
26 22.824
27 22.575
28 24.488
29 23.685
30 25.021
31 24.417
32 25.264
33 24.065
34 23.424
35 22.467
36 25.046
37 25.95
38 25.685
39 25.764
40 22.68
41 22.198
42 16.864
43 15.705
44 12.705
45 12.152
46 10.774
47 12.985
48 11.19
49 10.983
50 10.048
51 11.274
52 10.425
53 10.604
54 12.221
55 11.567
56 14.203
57 13.183
58 15.082
59 21.46
60 20.885
61 21.259
62 21.916
63 21.162
64 21.523
65 21.958
66 15.079
67 14.594
68 13.535
69 13.097
70 13.302
71 14.936
72 17.305
73 15.709
74 16.739
75 14.19
76 13.175
77 14.653
78 11.458
79 13.072
80 12.139
81 11.241
82 11.467
83 12.831
84 11.768
85 11.514
86 11.986
87 10.315
88 12.041
89 12.697
90 16.373
91 14.757
92 15.813
93 17.926
94 16.869
95 16.266
96 14.089
97 18.605
98 18.047
99 19.073
100 18.732
101 13.779
102 17.222
103 17.431
104 17.524
105 19.68
106 20.072
107 21.042
108 22.757
109 23.863
110 22.796
111 20.766
112 22.598
113 17.242
114 17.125
115 12.644
116 11.874
117 13.046
118 10.606
119 10.08
120 10.328
121 10.9
122 10.763
123 10.156
124 10.292
125 10.367
126 10.648
127 10.111
128 10.195
129 12.187
130 11.609
131 10.095
132 10.069
133 10.386
134 10.08
135 10.149
136 10.697
137 11.236
138 12.684
139 12.562
140 14.914
141 16.651
142 16.825
143 16.875
144 15.943
145 15.009
146 15.485
147 15.735
148 16.972
149 17.548
150 15.293
151 19.262
152 17.089
153 15.932
154 18.223
155 17.488
156 16.922
157 17.624
158 15.716
159 13.906
160 16.525
161 17.316
162 18.21
163 19.107
164 20.176
165 22.58
166 26.717
167 26.389
168 25.98
169 24.07
170 23.76
171 19.364
172 20.981
173 20.693
174 20.919
175 22.729
176 23.42
177 25.336
178 30.483
179 25.629
180 27.194
181 25.34
182 23.786
183 28.089
184 25.555
185 25.428
186 27.625
187 29.33
188 25.497
189 23.161
190 28.499
191 25.809
192 19.991
193 21.796
194 27.977
195 23.797
196 24.159
197 26.625
198 24.154
199 23.522
200 25.705
201 30.059
202 29.507
203 29.861
204 34.598
205 29.555
206 33.151
207 37.41
208 28.85
209 26.866
210 33.427
211 30.256
212 29.348
213 32.162
214 30.218
215 34.294
216 32.567
217 32.161
218 28.684
219 30.026
220 30.398
221 29.057
222 25.907
223 23.664
224 20.076
225 17.118
226 14.22
227 12.703
228 13.79
229 12.186
230 12.397
231 10.111
232 10.217
233 10.3
234 10.1
235 10.047
236 10.013
237 10.001
238 10.049
239 10.288
240 10.165
241 10.22
242 10.213
243 10.394
244 10.004
245 10.091
246 10.495
247 10.397
248 10.044
249 10.024
250 10.01
251 10.246
252 10.007
253 10.143
254 10.375
255 10.47
256 12.493
257 15.089
258 14.49
259 20.162
260 20.183
261 13.237
262 12.357
263 14.496
264 18.328
265 18.82
266 22.204
267 23.089
268 21.368
269 18.895
270 17.723
271 18.872
272 18.395
273 16.693
274 18.49
275 17.515
276 13.664
277 11.023
278 9.958
279 10.08
280 9.967
281 10.005
282 9.977
283 9.968
284 9.946
285 10.074
286 10.219
287 10.198
288 10.162
289 10.255
290 10.135
291 11.659
292 9.973
293 10.137
294 10.072
295 11.069
296 11.815
297 10.939
298 10.124
299 10.039
300 10.128
301 11.367
302 11.688
303 10.259
304 10.214
305 12.426
306 10.302
307 10.182
308 14.336
309 13.542
310 17.731
311 17.208
312 16.607
313 16.685
314 18.109
315 16.35
316 15.979
317 14.324
318 16.567
319 15.375
320 15.915
321 18.267
322 18.273
323 20.969
324 18.988
325 21.253
326 22.013
327 20.278
328 21.559
329 22.362
330 25.978
331 27.443
332 32.85
333 30.603
334 30.706
335 31.301
336 28.834
337 26.158
338 21.097
339 27.658
340 25.515
341 25.692
342 28.849
343 29.184
344 26.065
345 30.258
346 31.639
347 33.213
348 28.913
349 30.203
350 26.794
351 29.14
352 27.435
353 24.769
354 26.61
};
\addlegendentry{S2D}
\end{axis}

\end{tikzpicture}

%% file: figures/depth_70m.tex
\begin{tikzpicture}

\definecolor{color0}{rgb}{0,0.396078431372549,0.741176470588235}
\definecolor{color1}{rgb}{0.890196078431372,0.447058823529412,0.133333333333333}
\definecolor{color2}{rgb}{0.635294117647059,0.67843137254902,0}

\begin{axis}[
legend cell align={left},
legend style={fill opacity=0.8, draw opacity=1, text opacity=1, at={(0.03,0.97)}, anchor=north west, draw=white!80!black},
tick align=outside,
tick pos=left,
width=8.8cm,
x grid style={white!69.0196078431373!black},
xlabel={frame number},
xmin=0, xmax=208,
xtick style={color=black},
y grid style={white!69.0196078431373!black},
ylabel={depth value in m},
ymin=7.33355, ymax=119.71145,
ytick style={color=black}
]
\addplot [ultra thick, white!50.1960784313725!black, dash pattern=on 1pt off 3pt on 3pt off 3pt]
table {%
0 75.93564
1 76.50945
2 76.90317
3 76.94351
4 77.05595
5 77.16759
6 77.29926
7 76.8736
8 76.47269
9 75.12306
10 73.8941
11 70.48599
12 66.64518
13 62.54411
14 57.68766
15 52.75183
16 48.62586
17 45.69081
18 42.93712
19 41.35227
20 39.78791
21 38.7063
22 38.27602
23 38.80803
24 39.88222
25 42.10189
26 44.68322
27 47.66589
28 51.0746
29 54.42892
30 57.6211
31 60.76163
32 63.0701
33 64.40757
34 64.41567
35 62.93255
36 60.36387
37 57.32271
38 54.55605
39 51.42069
40 48.56833
41 45.76883
42 43.76245
43 42.76159
44 42.02743
45 41.24102
46 40.35611
47 39.2079
48 38.50597
49 38.02754
50 38.01215
51 38.41289
52 39.25156
53 40.57243
54 41.54719
55 42.73244
56 43.38665
57 43.81621
58 43.86393
59 43.74647
60 43.41525
61 43.70727
62 44.46606
63 45.66806
64 47.24177
65 48.03146
66 49.28795
67 50.43716
68 51.31687
69 52.4568
70 52.59986
71 53.429
72 53.57944
73 53.31543
74 53.39802
75 54.12533
76 55.91909
77 58.82031
78 61.69371
79 64.56712
80 67.14066
81 68.83158
82 69.01255
83 67.18688
84 63.98556
85 60.63897
86 56.5454
87 51.65508
88 45.46947
89 39.97701
90 35.93006
91 32.8097
92 29.34268
93 26.52538
94 25.11702
95 24.81311
96 25.62245
97 26.95993
98 28.14244
99 29.45207
100 30.37714
101 31.24426
102 32.04383
103 32.35156
104 32.54437
105 33.67882
106 36.20858
107 38.78502
108 41.0569
109 43.2081
110 44.78874
111 46.1856
112 47.19197
113 48.25547
114 48.77812
115 49.5438
116 50.08301
117 51.12627
118 52.0582
119 53.46434
120 55.50243
121 56.82703
122 58.41703
123 60.06131
124 61.18422
125 62.14627
126 63.1538
127 63.54498
128 64.3633
129 65.18355
130 66.521
131 67.50432
132 68.24604
133 68.75249
134 69.25535
135 70.15638
136 70.55045
137 70.58053
138 71.08888
139 72.06555
140 72.58147
141 73.2077
142 73.75916
143 74.40697
144 75.4003
145 76.29326
146 77.4459
147 79.4269
148 80.33373
149 82.31247
150 83.094
151 83.43289
152 84.42017
153 84.92262
154 85.03133
155 85.40533
156 85.95497
157 85.26902
158 85.65386
159 85.79443
160 85.72707
161 86.28609
162 86.21999
163 86.1055
164 86.49685
165 86.54689
166 87.13465
167 88.02847
168 88.53402
169 89.52887
170 90.70893
171 91.86478
172 92.51792
173 93.58418
174 95.61358
175 95.6761
176 95.24822
177 95.25772
178 95.61382
179 95.44798
180 94.73516
181 94.03368
182 94.72377
183 94.71046
184 94.94868
185 95.06355
186 94.28531
187 93.42487
188 92.05338
189 89.79318
190 87.84781
191 84.51352
192 80.14795
193 75.04141
194 69.83456
195 65.01001
196 60.10058
197 55.99805
198 51.38801
199 48.30316
200 45.56761
201 43.41666
202 41.35985
203 39.26757
204 36.71765
205 34.45387
206 31.89836
207 29.17189
208 27.06482
};
\addlegendentry{GT}
\addplot [thick, color0, dash pattern=on 1pt off 1pt]
table {%
0 75.737
1 72.788
2 75.737
3 78.953
4 78.953
5 78.953
6 78.953
7 78.953
8 72.788
9 72.788
10 72.788
11 67.571
12 65.252
13 61.095
14 55.833
15 51.473
16 45.66
17 43.722
18 41.138
19 40.352
20 38.878
21 38.185
22 37.521
23 37.521
24 38.878
25 41.138
26 42.821
27 46.703
28 50.181
29 54.291
30 55.833
31 59.224
32 61.095
33 63.099
34 61.095
35 61.095
36 59.224
37 55.833
38 52.841
39 51.473
40 47.801
41 43.722
42 41.96
43 41.96
44 41.138
45 41.138
46 39.599
47 38.878
48 38.185
49 37.521
50 36.883
51 37.521
52 38.185
53 39.599
54 41.138
55 42.821
56 42.821
57 43.722
58 43.722
59 42.821
60 42.821
61 42.821
62 43.722
63 44.667
64 45.66
65 46.703
66 48.959
67 48.959
68 51.473
69 51.473
70 51.473
71 52.841
72 51.473
73 52.841
74 51.473
75 52.841
76 54.291
77 59.224
78 61.095
79 65.252
80 65.252
81 65.252
82 65.252
83 65.252
84 63.099
85 59.224
86 54.291
87 48.959
88 43.722
89 38.185
90 34.562
91 32.095
92 28.557
93 25.313
94 24.117
95 23.894
96 25.062
97 26.384
98 27.893
99 29.261
100 30.4
101 31.22
102 32.095
103 32.095
104 32.095
105 33.031
106 35.678
107 38.185
108 40.352
109 42.821
110 43.722
111 44.667
112 46.703
113 47.801
114 47.801
115 48.959
116 48.959
117 50.181
118 51.473
119 52.841
120 55.833
121 55.833
122 57.474
123 59.224
124 61.095
125 61.095
126 61.095
127 61.095
128 65.252
129 65.252
130 65.252
131 65.252
132 67.571
133 67.571
134 67.571
135 67.571
136 70.075
137 70.075
138 70.075
139 72.788
140 72.788
141 72.788
142 72.788
143 72.788
144 72.788
145 72.788
146 78.953
147 78.953
148 78.953
149 82.476
150 82.476
151 82.476
152 82.476
153 82.476
154 86.352
155 86.352
156 86.352
157 82.476
158 86.352
159 86.352
160 86.352
161 86.352
162 86.352
163 86.352
164 86.352
165 86.352
166 86.352
167 86.352
168 90.635
169 90.635
170 90.635
171 90.635
172 90.635
173 90.635
174 90.635
175 95.395
176 90.635
177 90.635
178 95.395
179 90.635
180 90.635
181 90.635
182 90.635
183 90.635
184 90.635
185 90.635
186 90.635
187 90.635
188 90.635
189 90.635
190 86.352
191 82.476
192 78.953
193 75.737
194 70.075
195 65.252
196 59.224
197 54.291
198 48.959
199 47.801
200 44.667
201 42.821
202 41.138
203 38.878
204 36.269
205 33.523
206 31.22
207 28.557
208 26.963
};
\addlegendentry{KH}
\addplot [thick, color1, dash pattern=on 5pt off 1pt]
table {%
0 77.872
1 80.416
2 82.914
3 79.542
4 82.749
5 80.816
6 80.314
7 79.639
8 77.546
9 78.281
10 72.977
11 68.135
12 66.461
13 60.994
14 54.928
15 55.076
16 54.95
17 56.837
18 58.228
19 52.762
20 52.741
21 48.659
22 52.979
23 52.265
24 52.234
25 56.367
26 59.09
27 59.408
28 61.383
29 70.076
30 74.45
31 72.438
32 78.51
33 75.08
34 80.87
35 72.82
36 73.559
37 63.88
38 61.38
39 61.067
40 61.972
41 60.69
42 60.854
43 55.987
44 57.346
45 55.384
46 52.598
47 48.936
48 47.622
49 45.391
50 51.846
51 48.724
52 49.345
53 49.984
54 54.88
55 55.242
56 61.879
57 56.806
58 56.396
59 57.727
60 58.49
61 62.935
62 69.474
63 58.179
64 56.844
65 53.814
66 61.341
67 57.589
68 63.12
69 68.218
70 69.38
71 69.221
72 68.34
73 67.659
74 66.861
75 72.934
76 73.982
77 71.478
78 71.792
79 81.208
80 82.317
81 93.426
82 88.278
83 78.386
84 76.121
85 69.033
86 63.649
87 64.485
88 57.467
89 54.93
90 50.48
91 46.238
92 41.62
93 40.99
94 39.786
95 38.283
96 35.494
97 39.715
98 59.413
99 103.364
100 103.883
101 51.12
102 52.921
103 59.444
104 104.967
105 52.65
106 52.462
107 50.598
108 49.803
109 50.978
110 50.958
111 56.693
112 62.027
113 60.541
114 59.778
115 58.643
116 60.37
117 58.568
118 59.716
119 65.113
120 63.087
121 70.466
122 69.29
123 69.001
124 69.734
125 72.194
126 72.674
127 70.248
128 71.157
129 70.375
130 68.746
131 72.495
132 72.86
133 69.136
134 73.008
135 71.243
136 77.91
137 76.628
138 73.813
139 68.965
140 75.519
141 72.097
142 75.463
143 71.969
144 71.395
145 75.358
146 77.397
147 76.046
148 79.1
149 83.958
150 83.061
151 87.984
152 86.892
153 94.996
154 90.532
155 93.416
156 91.289
157 94.648
158 92.711
159 91.739
160 89.165
161 82.347
162 85.622
163 88.429
164 90.056
165 90.989
166 97.128
167 103.506
168 99.644
169 99.967
170 100.19
171 100.914
172 96.323
173 103
174 104.967
175 104.967
176 104.108
177 103.965
178 104.077
179 104.002
180 103.871
181 99.962
182 101.699
183 104.378
184 101.608
185 102.96
186 104.823
187 104.427
188 103.87
189 99.962
190 95.427
191 97.169
192 87.501
193 93.927
194 95.436
195 93.758
196 81.238
197 74.873
198 67.433
199 63.666
200 58.71
201 57.759
202 54.712
203 51.439
204 47.824
205 45.548
206 41.801
207 41.306
208 35.701
};
\addlegendentry{DD}
\addplot [thick, color2]
table {%
0 68.93
1 65.891
2 65.257
3 67.759
4 72.652
5 60.766
6 65.007
7 61.791
8 65.995
9 58.826
10 57.012
11 53.321
12 52.996
13 53.992
14 49.01
15 47.268
16 45.736
17 46.309
18 45.797
19 33.356
20 32.945
21 29.6
22 30.644
23 27.531
24 42.07
25 43.424
26 46.973
27 51.661
28 48.254
29 48.359
30 40.433
31 54.655
32 55.079
33 50.17
34 54.83
35 55.876
36 55.625
37 52.404
38 55.214
39 56.032
40 46.451
41 56.032
42 49.653
43 49.2
44 42.755
45 45.482
46 38.123
47 35.097
48 39.261
49 41.171
50 38.057
51 37.575
52 27.775
53 24.233
54 29.047
55 42.052
56 33.377
57 21.065
58 17.871
59 17.234
60 33.481
61 40.379
62 43.515
63 41.912
64 38.989
65 39.607
66 46.075
67 48.768
68 44.717
69 55.183
70 53.282
71 51.533
72 54.729
73 53.795
74 58.382
75 58.149
76 55.114
77 58.382
78 56.152
79 55.335
80 59.877
81 60.721
82 64.727
83 56.323
84 53.67
85 56.349
86 56.786
87 50.068
88 24.291
89 29.323
90 19.207
91 11.269
92 10.696
93 11.678
94 10.809
95 17.015
96 17.599
97 19.789
98 24.301
99 18.975
100 17.942
101 21.313
102 31.807
103 23.96
104 27.732
105 27.72
106 24.953
107 38.739
108 35.863
109 42.46
110 44.638
111 44.286
112 41.291
113 44.685
114 41.507
115 44.413
116 40.243
117 47.193
118 62.066
119 72.178
120 63.231
121 66.774
122 72.232
123 69.967
124 67.502
125 64.818
126 75.376
127 69.155
128 68.788
129 68.783
130 69.719
131 65.858
132 56.337
133 51.038
134 55.934
135 58.529
136 59.82
137 70.84
138 76.003
139 80.138
140 76.387
141 75.336
142 76.435
143 74.511
144 73.74
145 71.411
146 74.226
147 75.223
148 83.89
149 80.446
150 76.578
151 75.9
152 85.613
153 85.512
154 91.317
155 85.524
156 84.674
157 83.475
158 73.793
159 72.826
160 78.672
161 74.272
162 67.178
163 71.689
164 65.91
165 72.234
166 73.566
167 73.605
168 72.574
169 75.53
170 88.96
171 77.536
172 84.863
173 92.84
174 98.596
175 86.019
176 77.488
177 80.477
178 72.535
179 22.684
180 72.728
181 80.888
182 78.613
183 81.65
184 70.716
185 79.646
186 73.842
187 78.091
188 78.401
189 80.49
190 60.979
191 52.548
192 62.012
193 71.978
194 61.7
195 48.585
196 47.4
197 45.823
198 46.641
199 48.008
200 45.146
201 45.593
202 36.871
203 16.562
204 11.286
205 10.196
206 10.995
207 10.078
208 10.297
};
\addlegendentry{S2D}
\end{axis}

\end{tikzpicture}

%% file: figures/depth_rw.tex
\begin{tikzpicture}

\definecolor{color0}{rgb}{0.890196078431372,0.447058823529412,0.133333333333333}
\definecolor{color1}{rgb}{0.635294117647059,0.67843137254902,0}

\begin{axis}[
legend cell align={left},
legend style={fill opacity=0.8, draw opacity=1, text opacity=1, at={(0.03,0.97)}, anchor=north west, draw=white!80!black},
tick align=outside,
tick pos=left,
width=8.8cm,
x grid style={white!69.0196078431373!black},
xlabel={frame number},
xmin=0, xmax=738,
xtick style={color=black},
y grid style={white!69.0196078431373!black},
ylabel={depth value in m},
ymin=7.83685, ymax=133.72415,
ytick style={color=black}
]
\addplot [ultra thick, white!50.1960784313725!black, dash pattern=on 1pt off 3pt on 3pt off 3pt]
table {%
0 46.1819970388455
1 48.6659911165364
2 50.5290014805773
3 50.8029230769231
4 50.1087692307692
5 49.5881538461538
6 49.0675384615385
7 49.973997854235
8 51.05399570847
9 52.133993562705
10 53.247
11 53.346
12 53.445
13 53.577
14 53.676
15 53.6275003412359
16 53.1695017061795
17 52.826
18 52.57025
19 52.22925
20 52.11125
21 52.062125
22 51.996625
23 51.9475
24 51.898375
25 52.3578773636919
26 52.8337509454768
27 52.9941111111111
28 52.3665555555556
29 51.8958888888889
30 52.01375
31 52.56325
32 52.7569983102117
33 52.5139990344067
34 52.19
35 51.830375
36 51.47075
37 50.99125
38 50.631625
39 50.272
40 49.5867469366954
41 49.0728102025216
42 48.5588734683477
43 47.8736244894492
44 47.9321111111111
45 49.1354444444444
46 50.7398888888889
47 51.2137506504344
48 51.3228753252172
49 51.4805
50 51.626
51 51.7715
52 51.902875
53 51.98575
54 51.9620001177191
55 51.6459987050898
56 51.4089990582471
57 51.1719994114044
58 50.8559998822809
59 50.7236
60 50.6435
61 50.5367
62 50.2466640506877
63 49.8516648354814
64 49.3249992152063
65 48.93
66 48.5307
67 47.9983
68 47.599
69 47.8768456442798
70 48.2473087114404
71 48.5251543557202
72 48.803
73 49.3106
74 49.6913
75 50.072
76 49.5130011106321
77 49.0937486117099
78 48.674499444684
79 48.4642
80 48.568
81 48.6718
82 48.5840003743166
83 48.3485009357916
84 48.1130014972665
85 48.0931666666667
86 48.2989166666667
87 48.5046666666667
88 48.779
89 48.77625
90 48.7735
91 48.7698333333333
92 48.702918510437
93 48.5076786966324
94 48.2473589448929
95 48.0521191310883
96 47.8568793172836
97 47.5965595655441
98 47.4013197517395
99 47.2060799379349
100 46.8457482401747
101 46.5504988267831
102 46.2552494133916
103 46.16675
104 46.787
105 47.40725
106 48.23425
107 48.2986652527135
108 48.0851656768995
109 47.8004995758141
110 47.587
111 47.5417857142857
112 47.4815
113 47.4362857142857
114 47.3910714285714
115 47.6432515929325
116 47.910501061955
117 48.1777505309775
118 48.5044166666667
119 48.6826666666667
120 48.8609166666667
121 49.0985833333333
122 49.0049985408804
123 48.7754990880502
124 48.46949981761
125 48.4015454545455
126 48.4143636363636
127 48.4314545454545
128 48.4512999730587
129 48.485199892235
130 48.5304000538825
131 48.7442
132 49.3178
133 50.0826
134 50.3498
135 50.0042
136 49.5434
137 49.1888973371067
138 48.8165982247378
139 48.3201994082459
140 48.0306
141 47.9064
142 47.7408
143 47.6930999163152
144 47.7983996652608
145 47.9388001673696
146 47.88575
147 47.516
148 47.023
149 46.65325
150 47.557397550491
151 49.6121926514731
152 51.1533012247545
153 51.4122002429962
154 50.9026007289886
155 50.520401093483
156 50.1382014579773
157 49.6286019439697
158 49.246402308464
159 48.8642026729584
160 48.77875
161 49.669
162 50.55925
163 51.74625
164 51.7253339644796
165 51.2488349111991
166 50.6135028401584
167 50.137
168 50.2830769230769
169 50.4778461538462
170 50.6239230769231
171 50.77
172 50.7544
173 50.7427
174 50.731
175 51.1703350790853
176 51.4998344244283
177 51.8293337697713
178 51.6931666666667
179 51.1594166666667
180 50.6256666666667
181 49.914
182 50.3285798376736
183 50.7431563805858
184 51.2959251044686
185 51.7105016473807
186 52.1250781902929
187 52.6778469141757
188 53.0924234570879
189 53.507
190 53.4245381212632
191 53.3626920808422
192 53.3008460404211
193 53.2404999974455
194 53.2449999897821
195 53.2494999821186
196 53.2554999719007
197 53.26
198 53.1314285714286
199 52.96
200 52.8314285714286
201 52.7028571428571
202 52.4129289921884
203 52.1658579843768
204 51.9187869765652
205 51.5893570026039
206 51.6052666666667
207 51.7526666666667
208 51.9492
209 52.0966
210 52.244
211 52.0940002554483
212 51.9814995529655
213 51.8689997445517
214 51.719
215 51.2756
216 50.8322
217 50.241
218 49.7976
219 49.6280830828285
220 50.1324184202004
221 50.5106676686859
222 50.8889169171715
223 51.5667692307692
224 52.1185384615385
225 52.6703076923077
226 53.406
227 53.37155975914
228 53.3371197919846
229 53.2911998357773
230 53.2567598686218
231 53.2223199014664
232 53.1763999452591
233 53.1419599781036
234 53.0180835338365
235 52.6144152631443
236 52.3116658646539
237 52.0089164661635
238 51.9051200027466
239 51.9022400054932
240 51.8993600082397
241 51.8955200119019
242 51.8926399917603
243 51.8897599945068
244 51.8859199981689
245 51.7835
246 51.482
247 51.08
248 50.7785
249 51.0589990916265
250 51.821001816747
251 52.3925004541867
252 52.6002727272727
253 52.6348181818182
254 52.6607272727273
255 52.5736002489086
256 52.1560012445432
257 51.8427995021827
258 51.741
259 52.169
260 52.49
261 52.5641
262 52.0045
263 51.5848
264 51.3189999666214
265 51.3750001668928
266 51.4170000667571
267 51.4761818181818
268 51.6009090909091
269 51.6944545454545
270 51.788
271 52.053713380699
272 52.253
273 51.8226804103851
274 51.2489209575653
275 50.8186013679504
276 50.3882817783356
277 49.8145223255157
278 49.3842027359009
279 48.953883146286
280 49.1595
281 49.89825
282 50.637
283 50.3595
284 50.151375
285 50.1866663893952
286 50.3959991681855
287 50.553
288 50.85525
289 51.25825
290 51.38675
291 51.428375
292 51.44525
293 51.371
294 51.29675
295 51.6046266521629
296 51.9372506608652
297 52.1172941176471
298 51.9504705882353
299 51.8253529411765
300 51.7002352941176
301 51.5334117647059
302 51.448299963522
303 51.4431999756813
304 51.4363999918938
305 51.319
306 50.977
307 50.521
308 50.2147001866815
309 49.9798007467259
310 49.666599626637
311 49.5443636363636
312 49.6474545454545
313 49.7849090909091
314 49.888
315 49.6723005142681
316 49.384701199959
317 49.169
318 49.364
319 49.624
320 49.819
321 49.5012860749951
322 49.0776675083219
323 48.7599535833171
324 48.4422396583122
325 48.0186185666732
326 47.7009046416683
327 47.6626
328 47.7978
329 47.8992
330 47.9437999742508
331 47.9653999227525
332 47.9816000128746
333 47.737
334 47.237
335 46.862
336 46.4908181818182
337 45.9984545454545
338 45.6291818181818
339 45.5728995472438
340 46.3325022637812
341 46.9022009055125
342 47.2033
343 46.8885
344 46.6524
345 46.5178999454022
346 46.6094997270111
347 46.6782001091956
348 46.7073636363636
349 46.6408181818182
350 46.5909090909091
351 46.541
352 46.0585
353 45.696625
354 45.6752497042125
355 45.8737502957875
356 45.94375
357 45.856
358 45.739
359 45.9420621974135
360 46.1451243948269
361 46.4158739913782
362 46.6189378025865
363 46.822
364 46.3736
365 46.0373
366 45.701
367 46.0046
368 46.2323
369 46.46
370 45.9214525027418
371 45.517544287281
372 45.1136360718203
373 45.3999
374 45.8208
375 46.2417
376 46.1951004456037
377 46.0081994058618
378 45.8212998514654
379 45.4797
380 45.2004
381 44.9211
382 44.5991
383 44.3702
384 44.1413
385 43.7096644702741
386 43.3543317644815
387 42.9989990586889
388 42.5252221176321
389 42.1698894118395
390 41.8145567060469
391 41.3407797649901
392 40.9854442352642
393 41.28225
394 42.11275
395 42.735625
396 43.3585
397 44.189
398 43.7918009469972
399 43.3946018939945
400 42.865
401 42.8251
402 42.7852
403 42.732
404 43.3926
405 44.0532
406 44.934
407 44.8676246703125
408 44.801249868125
409 45.1165
410 45.65575
411 46.195
412 45.7847777777778
413 45.4771111111111
414 45.3435711848012
415 45.6298576303976
416 45.849125
417 46.0775
418 46.382
419 46.5492497507782
420 46.7164995015564
421 46.9394991692607
422 47.1067502492218
423 47.274
424 46.8249090909091
425 46.4880909090909
426 46.1512727272727
427 46.4791
428 46.9192
429 47.3593
430 47.4234995410449
431 47.3409997377399
432 47.258499934435
433 47.5343
434 47.8376
435 48.1409
436 48.2012000972746
437 48.1603998703005
438 48.1195999675751
439 48.40525
440 48.7045
441 49.00375
442 49.220000164906
443 48.9710006596241
444 48.7220011543421
445 48.389999835094
446 48.1676666666667
447 47.9586666666667
448 47.68
449 47.471
450 48.0569988357238
451 48.8383306166889
452 49.4243341095175
453 49.7937916877353
454 49.7089584386763
455 49.645333501882
456 49.5817085650878
457 49.4968753160288
458 49.4332503792345
459 49.3696254424403
460 49.4191538461538
461 49.7586153846154
462 50.0980769230769
463 50.5506923076923
464 50.8253332373037
465 50.9703329492149
466 51.1636658984298
467 51.3086667626963
468 51.6415
469 52.2105
470 52.63725
471 53.064
472 52.9266669395232
473 52.8236671441656
474 52.720667348808
475 52.6873333333333
476 52.7403333333333
477 52.7933333333333
478 52.864
479 52.6485
480 52.433
481 52.1456666666667
482 52.227421593151
483 52.9036702495644
484 53.8053351247822
485 54.4815837811956
486 54.453
487 53.945
488 53.564
489 53.183
490 53.2965389300517
491 53.3816926200345
492 53.4668463100172
493 53.68575
494 54.087
495 54.48825
496 55.02325
497 55.0429988675129
498 54.871999207259
499 54.6439996602539
500 54.473
501 54.62375
502 54.82475
503 54.9755
504 55.1835831195847
505 55.6139181629073
506 55.9366675216613
507 56.2594168804153
508 56.1792857142857
509 55.9915714285714
510 55.8038571428571
511 55.5535714285714
512 55.589429577165
513 55.7862863847766
514 55.9339289904854
515 56.0815715961942
516 55.5687153267269
517 54.6517883168172
518 53.7348613069075
519 52.8179342969978
520 52.2735
521 53.391
522 54.881
523 55.9985
524 56.2098336535424
525 55.8875009606272
526 55.6457514409408
527 55.404
528 55.6736
529 55.8758
530 56.078
531 56.2802
532 56.3280001728533
533 56.1975004321332
534 56.0235007778398
535 55.893
536 55.584
537 55.172
538 54.863
539 54.8440829616325
540 55.4053318465301
541 56.1536636930602
542 56.7149170383675
543 56.45025
544 55.9985
545 55.54675
546 55.0371667815711
547 54.8636671262842
548 54.6901674709974
549 54.4588345972816
550 54.5954285714286
551 54.8870714285714
552 55.2759285714286
553 55.5675714285714
554 55.7931537890181
555 55.9177689450904
556 56.0112303121447
557 56.1046924219638
558 56.2244285714286
559 56.3105714285714
560 56.3967142857143
561 56.5115714285714
562 56.5075001047338
563 56.3230004189352
564 56.0770008378704
565 55.8924996857986
566 55.708
567 55.7083333333333
568 55.7085833333333
569 55.7088333333333
570 55.7977141346349
571 55.9307853365872
572 56.0638575960954
573 56.2412858653651
574 56.1334166666667
575 55.5436666666667
576 54.7573333333333
577 54.1675833333333
578 54.1895329859842
579 54.6265989579525
580 54.9544010420475
581 55.2822005210237
582 55.6020833333333
583 55.5783333333333
584 55.5545833333333
585 55.5229166666667
586 55.8358326958954
587 56.317085564366
588 56.9587509561569
589 57.44
590 57.24
591 56.9733333333333
592 56.7733333333333
593 56.5725385852621
594 56.3026929263104
595 56.1003070736896
596 55.8979228294758
597 55.7728333333333
598 55.7875833333333
599 55.8023333333333
600 55.822
601 55.8987498475116
602 55.9755003049768
603 56.0778334349923
604 55.9420769230769
605 55.3813076923077
606 54.6336153846154
607 54.0728461538462
608 53.8664612313392
609 54.5363061566959
610 55.0386938433041
611 55.5410775373216
612 55.7568571428571
613 55.5781428571429
614 55.3994285714286
615 55.1611428571429
616 55.1316921431977
617 55.4007707112203
618 55.7595392840113
619 56.0286157136045
620 56.23525
621 56.34425
622 56.426
623 56.50775
624 56.1632507386001
625 55.7914985227997
626 55.4197492613999
627 55.0306666666667
628 54.9786666666667
629 54.9266666666667
630 54.8573333333333
631 55.1548327078163
632 55.6270855226428
633 56.2567509382755
634 56.729
635 56.3132636795756
636 55.7589485856763
637 55.3432122652519
638 54.9274759448274
639 54.3731608509282
640 54.2978888888889
641 54.9035555555555
642 55.7111111111111
643 56.3507777777778
644 57.0074444444444
645 57.883
646 58.5782479279977
647 59.2735013813348
648 60.0032
649 60.4025
650 60.8018
651 60.9769413041771
652 60.8403532604427
653 60.7037652167084
654 60.5216478250625
655 60.3850586958229
656 60.3452222222222
657 60.5501111111111
658 60.7037777777778
659 60.8376666666667
660 61.003
661 61.127
662 60.5851223084832
663 59.8626244616966
664 59.702
665 59.732
666 59.772
667 59.8446666666667
668 59.9173333333333
669 60.1205996886258
670 60.5124018682453
671 60.9042009341227
672 61.4482
673 61.9048
674 62.3614
675 62.6350913055352
676 62.0863652221408
677 61.5376347778592
678 60.806
679 61.3499
680 61.8938
681 62.619
682 62.8206
683 63.0222
684 63.291
685 63.2583331602594
686 63.2256665801297
687 63.349875
688 63.8205
689 64.291125
690 64.2556473285911
691 64.0632946571823
692 63.8709404570903
693 63.6144702285451
694 63.4221175571363
695 63.0955297798696
696 62.5705893396089
697 62.1768840094133
698 61.7831786792178
699 61.258238238957
700 61.8604
701 62.9605
702 64.4273
703 64.8322
704 64.8895
705 64.9659
706 64.7682731970159
707 64.4431804089524
708 64.0097268029841
709 64.0048
710 64.6402
711 65.4874
712 65.7745003254409
713 65.3650013017636
714 64.8190026035272
715 64.6001818181818
716 64.7627272727273
717 64.9794545454545
718 65.142
719 65.1189
720 65.0881
721 65.065
722 64.6875432729196
723 64.1842716364598
724 63.8068179091149
725 63.8604
726 64.2192
727 64.4883
728 64.6411998493197
729 64.767599547959
730 64.8624000753402
731 64.7905454545455
732 64.5836363636364
733 64.4284545454545
734 64.325
735 64.325
736 64.325
737 64.325
};
\addlegendentry{GT}
\addplot [thick, color0, dash pattern=on 5pt off 1pt]
table {%
0 49.291
1 47.695
2 48.101
3 48.324
4 50.793
5 49.76
6 50.027
7 49.657
8 50.158
9 47.504
10 49.302
11 48.318
12 51.465
13 49.005
14 49.222
15 44.934
16 45.441
17 44.324
18 45.451
19 44.021
20 45.375
21 45.535
22 46.773
23 47.79
24 49.401
25 48.51
26 46.996
27 46.35
28 46.507
29 45.802
30 46.021
31 45.661
32 44.563
33 45.714
34 43.456
35 40.385
36 41.506
37 42.662
38 41.134
39 44.843
40 44.307
41 44.285
42 45.013
43 45.12
44 40.808
45 41.86
46 41.111
47 46.014
48 45.269
49 41.899
50 44.457
51 40.669
52 38.07
53 39.087
54 41.397
55 41.955
56 43.553
57 48.17
58 51.787
59 48.435
60 42.023
61 43.963
62 39.492
63 44.281
64 41.963
65 41.944
66 37.854
67 42.167
68 42.683
69 46.183
70 40.949
71 37.355
72 38.771
73 39.185
74 38.343
75 40.787
76 40.878
77 40.71
78 36.561
79 36.048
80 35.324
81 38.097
82 38.978
83 38.133
84 39.255
85 43.387
86 42.677
87 41.446
88 44.944
89 43.079
90 39.702
91 38.773
92 40.907
93 40.447
94 41.215
95 42.574
96 40.969
97 41.283
98 44.85
99 43.943
100 41.445
101 43.876
102 61.461
103 67.341
104 72.845
105 72.891
106 45.002
107 39.58
108 38.736
109 39.281
110 37.227
111 37.847
112 37.768
113 39.009
114 37.192
115 36.808
116 37.367
117 36.731
118 38.242
119 39.452
120 37.744
121 37.282
122 35.443
123 36.272
124 38.939
125 37.814
126 37.95
127 38.962
128 37.504
129 38.69
130 39.316
131 38.294
132 37.527
133 37.765
134 37.912
135 37.774
136 38.007
137 38.342
138 36.984
139 36.482
140 35.597
141 35.753
142 38.494
143 40.917
144 40.157
145 38.019
146 38.523
147 39.26
148 41.134
149 40.011
150 38.413
151 38.36
152 38.878
153 38.87
154 45.099
155 46.174
156 44.083
157 43.599
158 42.262
159 43.133
160 43.634
161 39.993
162 40.958
163 42.442
164 38.86
165 39.31
166 38.076
167 40.348
168 42.169
169 43.765
170 42.644
171 40.889
172 41.553
173 41.526
174 42.174
175 41.443
176 42.356
177 41.585
178 42.93
179 43.21
180 45.142
181 44.74
182 44.934
183 47.101
184 44.044
185 44.638
186 49.478
187 46.792
188 46.316
189 47.039
190 48.56
191 52.349
192 55.734
193 55.93
194 54.559
195 50.597
196 53.311
197 52.861
198 50.394
199 49.468
200 50.205
201 51.244
202 51.336
203 48.241
204 44.678
205 44.758
206 45.222
207 45.55
208 45.266
209 46.445
210 45.529
211 45.699
212 46.941
213 48.733
214 47.332
215 45.447
216 45.247
217 45.35
218 44.213
219 48.028
220 49.731
221 48.554
222 46.423
223 46.537
224 45.262
225 46.157
226 46.138
227 45.736
228 45.982
229 47.011
230 47.173
231 44.946
232 45.774
233 45.927
234 47.315
235 48.61
236 49.725
237 46.991
238 49.241
239 48.514
240 49.333
241 48.196
242 48.555
243 49.325
244 48.83
245 47.859
246 48.465
247 46.046
248 47.566
249 47.647
250 47.249
251 47.739
252 44.408
253 43.726
254 43.435
255 45.317
256 42.887
257 41.033
258 38.481
259 42.065
260 37.098
261 36.748
262 35.493
263 35.81
264 34.358
265 34.406
266 34.668
267 33.669
268 36.804
269 35.811
270 36.272
271 35.018
272 34.883
273 34.414
274 34.167
275 35.072
276 36.168
277 37.036
278 36.577
279 35.122
280 36.069
281 37.108
282 37.908
283 37.117
284 37.903
285 36.372
286 36.12
287 39.035
288 36.268
289 36.18
290 34.876
291 33.149
292 33.238
293 33.626
294 32.812
295 34.235
296 33.711
297 34.844
298 34.997
299 34.978
300 33.965
301 32.377
302 31.477
303 31.497
304 30.563
305 29.542
306 29.722
307 29.878
308 30.904
309 30.901
310 31.198
311 31.444
312 30.416
313 31.5
314 30.99
315 28.963
316 28
317 27.173
318 23.294
319 24.833
320 23.661
321 23.725
322 22.05
323 22.659
324 24.657
325 24.151
326 23.806
327 24.676
328 22.279
329 21.509
330 18.862
331 19.173
332 18.549
333 18.164
334 17.925
335 18.325
336 17.944
337 18.082
338 17.811
339 17.483
340 17.6
341 17.583
342 17.672
343 17.477
344 17.576
345 17.431
346 16.323
347 16.346
348 16.47
349 17.512
350 17.307
351 17.092
352 16.915
353 16.476
354 16.358
355 16.184
356 15.634
357 16.116
358 16.207
359 16.492
360 16.753
361 16.79
362 16.722
363 16.52
364 16.785
365 16.494
366 16.052
367 16.044
368 15.704
369 16.19
370 17.621
371 17.092
372 17.039
373 17.015
374 16.994
375 16.787
376 15.648
377 16.76
378 16.838
379 17.565
380 17.174
381 16.966
382 15.91
383 15.736
384 16.388
385 16.366
386 15.469
387 15.393
388 15.61
389 15.593
390 15.331
391 17.36
392 16.478
393 15.918
394 17.368
395 15.869
396 16.147
397 15.912
398 16.324
399 15.571
400 15.59
401 15.648
402 15.049
403 15.511
404 15.931
405 15.484
406 15.376
407 16.329
408 16.411
409 15.704
410 15.89
411 16.494
412 16.093
413 16.037
414 16.564
415 14.959
416 16.083
417 15.271
418 14.858
419 15.944
420 18.56
421 20.237
422 24.258
423 30.555
424 36.85
425 33.205
426 35.194
427 36.136
428 37.083
429 42.107
430 49.913
431 49.147
432 52.373
433 59.185
434 54.9
435 57.801
436 57.594
437 58.937
438 72.478
439 73.574
440 69.484
441 63.006
442 62.955
443 55.325
444 54.053
445 52.141
446 56.846
447 59.204
448 57.511
449 52.499
450 55.351
451 52.048
452 51.09
453 50.418
454 45.172
455 42.299
456 41.629
457 40.317
458 41.936
459 38.686
460 38.347
461 39.771
462 40.057
463 39.138
464 38.684
465 40.766
466 39.555
467 37.667
468 38.57
469 37.564
470 38.031
471 37.364
472 37.556
473 38.509
474 37.378
475 35.534
476 36.217
477 34.055
478 35.122
479 37.084
480 34.669
481 36.085
482 36.101
483 35.732
484 37.469
485 37.74
486 37.947
487 36.881
488 38.696
489 39.496
490 40.588
491 40.397
492 40.386
493 38.947
494 39.167
495 37.295
496 36.427
497 40.996
498 41.775
499 42.653
500 42.903
501 45.335
502 41.373
503 39.864
504 40.298
505 39.912
506 41.219
507 40.788
508 39.956
509 39.398
510 40.734
511 42.379
512 39.835
513 38.8
514 39.306
515 39.464
516 39.552
517 39.886
518 38.896
519 38.416
520 38.248
521 38.224
522 38.032
523 38.113
524 37.531
525 35.66
526 36.162
527 36.132
528 36.076
529 36.814
530 38.16
531 36.732
532 35.943
533 36.785
534 37.22
535 34.935
536 37.262
537 35.832
538 37.84
539 37.927
540 36.934
541 36.723
542 36.917
543 37.31
544 37.293
545 36.694
546 36.856
547 36.595
548 37.004
549 36.319
550 35.665
551 36.037
552 35.029
553 35.838
554 34.895
555 35.932
556 37.479
557 37.025
558 37.392
559 36.914
560 36.659
561 35.653
562 35.552
563 35.799
564 35.786
565 35.45
566 34.981
567 36.512
568 36.24
569 37.937
570 36.562
571 36.952
572 35.341
573 37.122
574 35.363
575 35.252
576 34.767
577 34.059
578 34.351
579 35.981
580 37.587
581 34.579
582 37.823
583 38.903
584 40.433
585 39.291
586 37.846
587 38.428
588 39.394
589 40.23
590 38.907
591 39.787
592 40.496
593 40.804
594 39.396
595 40.582
596 39.225
597 38.451
598 38.065
599 39.347
600 39.154
601 38.735
602 39.781
603 41.459
604 39.292
605 40.372
606 39.589
607 39.156
608 38.408
609 39.641
610 40.14
611 41.227
612 39.78
613 40.218
614 39.946
615 41.159
616 40.368
617 39.982
618 41.111
619 42.051
620 40.023
621 40.103
622 39.207
623 40.715
624 41.094
625 42.235
626 41.622
627 42.1
628 41.544
629 41.25
630 40.888
631 39.948
632 39.357
633 41.854
634 39.332
635 39.342
636 38.299
637 38.079
638 40.182
639 39.034
640 39.696
641 39.042
642 38.473
643 36.838
644 38.367
645 39.551
646 38.189
647 38.998
648 41.279
649 41.152
650 39.989
651 42.093
652 39.128
653 38.728
654 42.056
655 41.612
656 40.876
657 39.769
658 44.747
659 42.708
660 41.283
661 41.493
662 41.552
663 46.977
664 44.03
665 44.143
666 45.668
667 47.737
668 45.794
669 46.193
670 44.574
671 46.349
672 45.335
673 47.478
674 43.619
675 44.82
676 45.994
677 46.666
678 47.229
679 44.425
680 47.363
681 43.633
682 42.134
683 43.857
684 43.329
685 45.411
686 45.95
687 46.167
688 47.12
689 47.572
690 45.775
691 44.429
692 46.618
693 47.246
694 46.817
695 48.242
696 46.119
697 47.712
698 48.555
699 45.408
700 45.243
701 44.951
702 45.579
703 49.059
704 47.012
705 47.257
706 46.258
707 44.727
708 44.196
709 43.617
710 43.719
711 43.732
712 43.555
713 43.388
714 43.02
715 45.415
716 45.056
717 46.665
718 46.503
719 47.263
720 46.348
721 48.279
722 45.005
723 45.787
724 46.455
725 44.308
726 45.977
727 45.008
728 45.061
729 45.805
730 44.235
731 46.82
732 44.807
733 45.505
734 43.767
735 43.558
736 43.399
737 42.694
};
\addlegendentry{DD}
\addplot [thick, color1]
table {%
0 38.52
1 44.074
2 42.595
3 46.388
4 49.115
5 51.017
6 47.599
7 46.398
8 44.944
9 44.749
10 33.845
11 37.88
12 41.241
13 30.957
14 38.672
15 34.381
16 33.294
17 34.102
18 34.658
19 32.15
20 36.064
21 34.38
22 33.586
23 29.544
24 29.199
25 28.859
26 30.265
27 31.726
28 31.539
29 30.782
30 41.85
31 48.363
32 45.342
33 39.631
34 39.8
35 59.951
36 42.298
37 49.002
38 36.119
39 31.907
40 39.838
41 40.765
42 37.052
43 40.228
44 40.418
45 39.227
46 44.478
47 47.611
48 40.761
49 45.794
50 45.366
51 44.774
52 60.753
53 58.616
54 58.381
55 45.771
56 48.193
57 44.786
58 41.405
59 43.777
60 40.36
61 43.526
62 44.208
63 55.1
64 51.012
65 60.861
66 41.88
67 52.702
68 49.13
69 41.728
70 49.422
71 44.241
72 44.786
73 43.644
74 44.538
75 34.097
76 35.233
77 35.972
78 56.86
79 32.778
80 39.427
81 34.125
82 28.954
83 36.337
84 32.571
85 35.18
86 49.339
87 35.965
88 38.374
89 42.625
90 31.921
91 38.108
92 43.932
93 35.944
94 38.211
95 37.252
96 41.845
97 35.583
98 42.175
99 39.114
100 46.738
101 63.084
102 34.156
103 50.174
104 50.607
105 47.793
106 51.578
107 64.14
108 73.857
109 72.252
110 75.022
111 77.781
112 76.276
113 75.937
114 69.217
115 71.436
116 79.847
117 73.736
118 75.299
119 73.848
120 73.89
121 73.902
122 77.158
123 81.683
124 80.925
125 84.965
126 83.998
127 80.345
128 80.745
129 84.649
130 83.465
131 87.636
132 88.463
133 86.378
134 86.71
135 84.028
136 76.934
137 78.707
138 81.648
139 81.011
140 81.524
141 84.769
142 83.92
143 85.223
144 84.157
145 80.438
146 75.826
147 79.787
148 86.863
149 88.259
150 86.092
151 84.18
152 87.399
153 92.729
154 90.065
155 91.809
156 87.18
157 90.393
158 87.27
159 85.683
160 86.603
161 84.677
162 87.904
163 98.87
164 99.171
165 104.682
166 105.446
167 111.161
168 121.725
169 108.946
170 106.882
171 104.616
172 109.791
173 112.544
174 112.747
175 112.346
176 111.373
177 110.489
178 114.078
179 106.432
180 115.624
181 111.101
182 114.168
183 112.353
184 106.726
185 112.789
186 108.402
187 105.018
188 99.547
189 98.211
190 106.824
191 107.762
192 105.708
193 102.849
194 109
195 116.415
196 110.851
197 98.021
198 69.523
199 82.706
200 90.838
201 102.144
202 115.305
203 116.323
204 115.903
205 121.089
206 128.002
207 114.119
208 108.307
209 102.721
210 101.195
211 110.146
212 100.5
213 96.475
214 92.577
215 102.079
216 110.472
217 117.721
218 112.122
219 100.974
220 110.81
221 117.893
222 111.787
223 94.365
224 101.39
225 103.301
226 76.252
227 89.7
228 96.766
229 106.929
230 105.702
231 81.796
232 85.848
233 76.091
234 78.295
235 86.906
236 94.972
237 94.715
238 80.723
239 87.736
240 89.528
241 85.912
242 86.602
243 76.307
244 67.966
245 97.113
246 71.888
247 63.445
248 71.609
249 82.276
250 78.66
251 82.616
252 89.078
253 88.898
254 83.725
255 85.977
256 79.961
257 76.189
258 74.303
259 71.366
260 83.931
261 82.488
262 77.047
263 75.281
264 73.456
265 65.232
266 53.424
267 50.765
268 57.735
269 66.034
270 68.93
271 53.467
272 66.119
273 68.349
274 66.673
275 64.236
276 51.73
277 67.247
278 49.653
279 41.468
280 49.197
281 47.113
282 57.612
283 46.372
284 43.782
285 40.852
286 38.404
287 40.443
288 42.774
289 44.94
290 42.236
291 46.462
292 58.182
293 63.623
294 61.003
295 66.76
296 70.794
297 65
298 41.452
299 46.619
300 52.886
301 57.545
302 59.198
303 53.743
304 48.894
305 44.695
306 34.15
307 40.855
308 34.213
309 47.249
310 35.056
311 35.165
312 28.839
313 28.143
314 34.75
315 32.952
316 31.829
317 26.898
318 25.993
319 25.138
320 25.265
321 26.105
322 27.03
323 25.723
324 26.527
325 25.939
326 27.399
327 26.516
328 26.318
329 25.271
330 25.353
331 25.668
332 27.522
333 25.066
334 27.493
335 26.93
336 26.284
337 25.315
338 26.435
339 27.05
340 27.666
341 28.873
342 28.495
343 28.283
344 27.796
345 27.865
346 26.425
347 25.287
348 27.599
349 27.654
350 28.454
351 31.416
352 30.902
353 29.135
354 30.113
355 32.649
356 35.28
357 30.135
358 28.343
359 29.346
360 30.364
361 29.855
362 31.743
363 35.346
364 33.828
365 32.331
366 31.998
367 31.362
368 31.493
369 32.792
370 31.074
371 31.522
372 34.994
373 34.287
374 37.421
375 35.179
376 36.427
377 30.496
378 34.679
379 35.792
380 34.816
381 38.335
382 39.295
383 36.262
384 34.734
385 25.7
386 31.313
387 30.664
388 26.657
389 24.406
390 29.403
391 29.974
392 22.228
393 16.766
394 26.32
395 25.677
396 19.755
397 21.336
398 22.607
399 23.509
400 17.258
401 25.29
402 15.693
403 13.559
404 14.531
405 22.207
406 30.273
407 28.243
408 28.121
409 23.758
410 22.791
411 30.631
412 24.476
413 21.881
414 29.075
415 30.222
416 22.596
417 23.187
418 20.405
419 22.179
420 20.903
421 19.404
422 21.369
423 23.535
424 26.538
425 28.248
426 31.708
427 26.055
428 21.582
429 27.656
430 28.656
431 44.258
432 43.973
433 38.782
434 23.663
435 30.549
436 30.732
437 40.771
438 29.891
439 30.842
440 29.756
441 35.424
442 34.018
443 32.935
444 28.372
445 30.325
446 40.642
447 50.63
448 56.453
449 51.945
450 50.148
451 49.694
452 31.966
453 53.586
454 55.612
455 53.553
456 47.068
457 49.696
458 57.657
459 50.894
460 44.251
461 43.28
462 45.631
463 54.906
464 32.925
465 44.279
466 58.074
467 73.077
468 67.207
469 54.482
470 62.536
471 70.459
472 59.552
473 62.723
474 59.743
475 52.952
476 75.743
477 76.168
478 72.82
479 83.02
480 72.132
481 82.626
482 68.073
483 69.372
484 83.211
485 88.333
486 88.817
487 82.266
488 89.861
489 94.306
490 87.615
491 88.889
492 84.481
493 85.504
494 91.366
495 90.291
496 88.749
497 87.004
498 81.149
499 76.634
500 80.045
501 86.23
502 91.267
503 92.983
504 86.886
505 91.77
506 94.32
507 93.094
508 87.319
509 85.998
510 90.424
511 91.866
512 93.365
513 87.726
514 85.699
515 85.295
516 86.39
517 88.781
518 83.355
519 82.844
520 81.776
521 85.669
522 88.322
523 89.989
524 90.998
525 92.69
526 91.468
527 90.353
528 87.701
529 84.963
530 88.69
531 87.043
532 82.77
533 73.981
534 79.926
535 83.003
536 70.364
537 80.792
538 77.647
539 80.708
540 79.535
541 83.385
542 88.176
543 88.81
544 85.136
545 79.125
546 77.009
547 79.323
548 74.416
549 76.447
550 74.667
551 79.856
552 72.276
553 71.923
554 71.075
555 66.958
556 74.679
557 76.828
558 75.375
559 72.133
560 76.368
561 87.886
562 85.39
563 85.45
564 79.841
565 82.275
566 79.769
567 77.283
568 65.077
569 68.006
570 70.814
571 67.773
572 70.541
573 67.018
574 78.194
575 88.872
576 86.606
577 82.619
578 76.195
579 80.714
580 78.647
581 75.245
582 69.922
583 85.219
584 82.761
585 85.779
586 87.69
587 77.59
588 71.026
589 72.233
590 72.374
591 88.855
592 77.612
593 76.992
594 84.43
595 87.666
596 76.645
597 81.161
598 80.196
599 83.396
600 78.72
601 88.707
602 88.721
603 87.304
604 87.548
605 79.886
606 83.155
607 81.272
608 80.647
609 83.109
610 78.607
611 81.167
612 84.873
613 88.513
614 83.378
615 82.071
616 87.013
617 67.144
618 79.598
619 91.563
620 87.595
621 86.156
622 85.019
623 89.261
624 89.942
625 88.131
626 86.341
627 82.59
628 85.345
629 92.407
630 94.455
631 89.775
632 89.982
633 79.944
634 92.97
635 93.186
636 85.778
637 84.72
638 86.128
639 84.171
640 86.338
641 79.985
642 81.754
643 81.79
644 84.122
645 81.682
646 83.628
647 76.192
648 82.391
649 84.405
650 86.575
651 88.425
652 90.822
653 89.315
654 97.281
655 94.737
656 96.787
657 98.6
658 100.426
659 96
660 94.234
661 90.891
662 82.294
663 86.276
664 80.897
665 85.086
666 82.798
667 71.618
668 75.301
669 80.61
670 90.473
671 88.247
672 76.277
673 85.287
674 80.156
675 87.931
676 86.185
677 96.605
678 95.621
679 101.45
680 102.915
681 98.975
682 95.289
683 98.711
684 75.15
685 75.964
686 81.242
687 65.093
688 40.804
689 63.974
690 70.174
691 83.523
692 85.703
693 96.33
694 103.067
695 96.935
696 90.492
697 98.049
698 88.079
699 97.145
700 102.231
701 95.918
702 90.644
703 80.231
704 81.845
705 69.116
706 66.004
707 71.36
708 48.568
709 59.134
710 64.832
711 61.802
712 70.904
713 72.441
714 86.531
715 92.047
716 86.283
717 80.025
718 85.384
719 81.219
720 85.18
721 80.15
722 64.123
723 67.337
724 65.614
725 61.791
726 67.084
727 64.252
728 63.026
729 58.721
730 53.98
731 47.355
732 46.022
733 62.265
734 47.625
735 60.557
736 33.541
737 44.92
};
\addlegendentry{S2D}
\addplot [semithick, black]
table {%
0 53.88
1 53.88
2 56.813
3 56.813
4 56.813
5 56.813
6 51.256
7 53.88
8 53.88
9 53.88
10 56.813
11 51.256
12 51.256
13 56.813
14 53.88
15 53.88
16 53.88
17 53.88
18 53.88
19 51.256
20 53.88
21 53.88
22 51.256
23 51.256
24 48.894
25 46.757
26 48.894
27 46.757
28 46.757
29 48.894
30 48.894
31 48.894
32 51.256
33 51.256
34 53.88
35 48.894
36 51.256
37 nan
38 46.757
39 51.256
40 48.894
41 48.894
42 51.256
43 51.256
44 51.256
45 48.894
46 48.894
47 46.757
48 46.757
49 46.757
50 46.757
51 48.894
52 48.894
53 46.757
54 48.894
55 46.757
56 48.894
57 48.894
58 nan
59 nan
60 48.894
61 51.256
62 51.256
63 46.757
64 46.757
65 48.894
66 48.894
67 48.894
68 48.894
69 48.894
70 51.256
71 48.894
72 48.894
73 46.757
74 46.757
75 46.757
76 46.757
77 44.814
78 46.757
79 46.757
80 48.894
81 46.757
82 46.757
83 46.757
84 46.757
85 46.757
86 46.757
87 46.757
88 46.757
89 46.757
90 48.894
91 46.757
92 46.757
93 46.757
94 46.757
95 46.757
96 48.894
97 51.256
98 51.256
99 51.256
100 48.894
101 48.894
102 46.757
103 48.894
104 48.894
105 48.894
106 46.757
107 48.894
108 48.894
109 48.894
110 48.894
111 48.894
112 51.256
113 48.894
114 48.894
115 48.894
116 48.894
117 48.894
118 48.894
119 48.894
120 48.894
121 48.894
122 46.757
123 48.894
124 48.894
125 48.894
126 51.256
127 48.894
128 51.256
129 48.894
130 51.256
131 48.894
132 48.894
133 51.256
134 51.256
135 48.894
136 48.894
137 51.256
138 48.894
139 48.894
140 48.894
141 48.894
142 48.894
143 48.894
144 48.894
145 48.894
146 48.894
147 48.894
148 51.256
149 51.256
150 48.894
151 51.256
152 51.256
153 48.894
154 48.894
155 51.256
156 51.256
157 53.88
158 51.256
159 53.88
160 51.256
161 51.256
162 51.256
163 53.88
164 51.256
165 51.256
166 51.256
167 51.256
168 51.256
169 53.88
170 53.88
171 53.88
172 53.88
173 51.256
174 51.256
175 51.256
176 48.894
177 51.256
178 53.88
179 51.256
180 53.88
181 53.88
182 51.256
183 56.813
184 53.88
185 53.88
186 51.256
187 51.256
188 51.256
189 51.256
190 51.256
191 53.88
192 51.256
193 51.256
194 51.256
195 53.88
196 53.88
197 53.88
198 48.894
199 51.256
200 53.88
201 53.88
202 51.256
203 51.256
204 51.256
205 53.88
206 51.256
207 51.256
208 51.256
209 53.88
210 51.256
211 51.256
212 51.256
213 51.256
214 48.894
215 51.256
216 51.256
217 51.256
218 51.256
219 51.256
220 51.256
221 51.256
222 51.256
223 51.256
224 51.256
225 51.256
226 51.256
227 51.256
228 53.88
229 51.256
230 53.88
231 53.88
232 53.88
233 48.894
234 51.256
235 51.256
236 51.256
237 51.256
238 53.88
239 51.256
240 51.256
241 48.894
242 51.256
243 48.894
244 48.894
245 48.894
246 51.256
247 48.894
248 48.894
249 48.894
250 48.894
251 48.894
252 51.256
253 48.894
254 48.894
255 48.894
256 48.894
257 46.757
258 44.814
259 46.757
260 46.757
261 48.894
262 48.894
263 46.757
264 48.894
265 46.757
266 46.757
267 48.894
268 48.894
269 48.894
270 46.757
271 46.757
272 48.894
273 48.894
274 48.894
275 48.894
276 48.894
277 48.894
278 51.256
279 46.757
280 48.894
281 48.894
282 48.894
283 46.757
284 48.894
285 48.894
286 46.757
287 48.894
288 46.757
289 44.814
290 46.757
291 48.894
292 48.894
293 48.894
294 48.894
295 48.894
296 46.757
297 46.757
298 46.757
299 48.894
300 48.894
301 48.894
302 51.256
303 48.894
304 48.894
305 51.256
306 51.256
307 51.256
308 48.894
309 51.256
310 51.256
311 48.894
312 51.256
313 51.256
314 51.256
315 53.88
316 51.256
317 51.256
318 51.256
319 51.256
320 51.256
321 51.256
322 51.256
323 51.256
324 51.256
325 51.256
326 51.256
327 48.894
328 51.256
329 48.894
330 51.256
331 51.256
332 48.894
333 51.256
334 51.256
335 53.88
336 51.256
337 46.757
338 48.894
339 48.894
340 48.894
341 48.894
342 48.894
343 48.894
344 46.757
345 46.757
346 46.757
347 46.757
348 46.757
349 48.894
350 48.894
351 48.894
352 46.757
353 48.894
354 46.757
355 46.757
356 46.757
357 46.757
358 46.757
359 48.894
360 48.894
361 48.894
362 46.757
363 48.894
364 48.894
365 46.757
366 48.894
367 51.256
368 48.894
369 48.894
370 48.894
371 48.894
372 48.894
373 46.757
374 48.894
375 48.894
376 48.894
377 48.894
378 51.256
379 46.757
380 48.894
381 48.894
382 48.894
383 48.894
384 48.894
385 51.256
386 51.256
387 53.88
388 51.256
389 53.88
390 53.88
391 53.88
392 53.88
393 53.88
394 56.813
395 56.813
396 56.813
397 56.813
398 56.813
399 53.88
400 56.813
401 53.88
402 56.813
403 60.112
404 56.813
405 56.813
406 56.813
407 56.813
408 60.112
409 56.813
410 60.112
411 60.112
412 60.112
413 63.852
414 63.852
415 56.813
416 63.852
417 63.852
418 60.112
419 60.112
420 60.112
421 60.112
422 56.813
423 60.112
424 56.813
425 56.813
426 53.88
427 56.813
428 53.88
429 53.88
430 51.256
431 48.894
432 51.256
433 48.894
434 53.88
435 51.256
436 51.256
437 48.894
438 51.256
439 48.894
440 48.894
441 53.88
442 51.256
443 48.894
444 48.894
445 46.757
446 48.894
447 48.894
448 46.757
449 46.757
450 46.757
451 46.757
452 46.757
453 48.894
454 46.757
455 48.894
456 48.894
457 46.757
458 48.894
459 51.256
460 48.894
461 51.256
462 46.757
463 48.894
464 51.256
465 51.256
466 48.894
467 48.894
468 51.256
469 51.256
470 51.256
471 51.256
472 53.88
473 51.256
474 53.88
475 53.88
476 53.88
477 53.88
478 53.88
479 51.256
480 51.256
481 51.256
482 53.88
483 56.813
484 53.88
485 53.88
486 53.88
487 53.88
488 51.256
489 56.813
490 51.256
491 51.256
492 51.256
493 53.88
494 51.256
495 53.88
496 51.256
497 56.813
498 56.813
499 56.813
500 60.112
501 53.88
502 56.813
503 56.813
504 56.813
505 56.813
506 53.88
507 56.813
508 56.813
509 56.813
510 53.88
511 56.813
512 56.813
513 56.813
514 53.88
515 53.88
516 56.813
517 56.813
518 56.813
519 56.813
520 56.813
521 60.112
522 60.112
523 56.813
524 60.112
525 56.813
526 56.813
527 56.813
528 56.813
529 60.112
530 60.112
531 60.112
532 56.813
533 60.112
534 56.813
535 56.813
536 60.112
537 56.813
538 56.813
539 56.813
540 56.813
541 56.813
542 56.813
543 56.813
544 56.813
545 56.813
546 56.813
547 56.813
548 53.88
549 56.813
550 56.813
551 60.112
552 60.112
553 56.813
554 56.813
555 60.112
556 60.112
557 56.813
558 56.813
559 60.112
560 60.112
561 60.112
562 60.112
563 60.112
564 60.112
565 60.112
566 60.112
567 60.112
568 56.813
569 60.112
570 56.813
571 60.112
572 60.112
573 60.112
574 56.813
575 56.813
576 60.112
577 60.112
578 60.112
579 60.112
580 60.112
581 56.813
582 56.813
583 56.813
584 56.813
585 56.813
586 56.813
587 60.112
588 56.813
589 56.813
590 56.813
591 56.813
592 56.813
593 56.813
594 60.112
595 56.813
596 56.813
597 60.112
598 56.813
599 56.813
600 56.813
601 56.813
602 56.813
603 53.88
604 56.813
605 56.813
606 56.813
607 56.813
608 56.813
609 56.813
610 56.813
611 56.813
612 56.813
613 56.813
614 56.813
615 56.813
616 56.813
617 56.813
618 56.813
619 56.813
620 56.813
621 56.813
622 56.813
623 56.813
624 56.813
625 60.112
626 60.112
627 60.112
628 60.112
629 60.112
630 60.112
631 60.112
632 60.112
633 63.852
634 60.112
635 60.112
636 56.813
637 56.813
638 56.813
639 60.112
640 60.112
641 56.813
642 56.813
643 56.813
644 60.112
645 60.112
646 60.112
647 60.112
648 60.112
649 63.852
650 60.112
651 56.813
652 60.112
653 60.112
654 60.112
655 60.112
656 56.813
657 60.112
658 60.112
659 60.112
660 60.112
661 60.112
662 60.112
663 60.112
664 60.112
665 60.112
666 60.112
667 60.112
668 60.112
669 56.813
670 60.112
671 60.112
672 60.112
673 60.112
674 60.112
675 60.112
676 60.112
677 60.112
678 56.813
679 63.852
680 56.813
681 60.112
682 60.112
683 63.852
684 60.112
685 63.852
686 63.852
687 60.112
688 60.112
689 60.112
690 56.813
691 60.112
692 60.112
693 60.112
694 60.112
695 60.112
696 63.852
697 60.112
698 56.813
699 60.112
700 60.112
701 63.852
702 63.852
703 60.112
704 63.852
705 63.852
706 60.112
707 63.852
708 63.852
709 60.112
710 60.112
711 60.112
712 63.852
713 60.112
714 60.112
715 63.852
716 60.112
717 60.112
718 63.852
719 60.112
720 60.112
721 60.112
722 60.112
723 60.112
724 60.112
725 60.112
726 60.112
727 60.112
728 56.813
729 56.813
730 60.112
731 56.813
732 56.813
733 60.112
734 56.813
735 56.813
736 60.112
737 56.813
};
\addlegendentry{SSD KH}
\end{axis}

\end{tikzpicture}